\pdfoutput=1

\documentclass[11pt]{article}

\usepackage[]{EMNLP2022}

\usepackage{times}
\usepackage{latexsym}

\usepackage{subfig}
\usepackage{amsmath}
\usepackage{float}
\usepackage{booktabs}
\usepackage{adjustbox}
\usepackage[]{graphicx}
\usepackage{subfig}
\usepackage{xcolor, soul}
\usepackage{multirow}
\usepackage[most]{tcolorbox}
\usepackage{tikz}
\usepackage{amssymb}
\usepackage{pifont}

\usepackage[T1]{fontenc}

\usepackage[utf8]{inputenc}

\usepackage{microtype}

\usepackage{inconsolata}
\usepackage{hyperref}

\newcommand{\cmark}{\ding{51}}

\newcommand\ti[1]{\textit{#1}}

\newcommand\tf[1]{\textbf{#1}}

\newcommand{\bertft}{BERT-NLI}

\renewcommand{\paragraph}[1]{\vspace{0.2cm}\noindent\textbf{#1}}

\newcommand{\ours}{MABEL}
\newcommand{\ff}{FairFil}

\definecolor{c1}{cmyk}{0,0.6175,0.8848,0.1490}
\definecolor{c2}{cmyk}{0.1127,0.6690,0,0.4431}
\definecolor{c3}{cmyk}{0.3081,0,0.7209,0.3255}
\definecolor{c4}{cmyk}{0.6765,0.2017,0,0.0667}
\definecolor{c5}{cmyk}{0,0.8765,0.7099,0.3647}

\newtcbox{\hlprimarytab}{on line, rounded corners, box align=base, colback=c3!10,colframe=white,size=fbox,arc=3pt, before upper=\strut, top=-2pt, bottom=-4pt, left=-2pt, right=-2pt, boxrule=0pt}
\newtcbox{\hlsecondarytab}{on line, box align=base, colback=red!10,colframe=white,size=fbox,arc=3pt, before upper=\strut, top=-2pt, bottom=-4pt, left=-2pt, right=-2pt, boxrule=0pt}

\newcommand{\uashifted}{\raisebox{0.5\depth}{\tiny$\uparrow$}}

\newcommand{\ua}[1]{{\small\hlsecondarytab{\uashifted{#1}}}}

\newcommand*{\TakeFourierOrnament}[1]{{%
\fontencoding{U}\fontfamily{futs}\selectfont\char#1}}
\newcommand*{\danger}{\TakeFourierOrnament{66}}

\title{{\ours}: Attenuating Gender Bias using Textual Entailment Data} 

\author{Jacqueline He\thanks{$\ \ $This work was done before JH graduated from Princeton University.} \quad Mengzhou Xia \quad Christiane Fellbaum \quad Danqi Chen\\
  \large{Department of Computer Science, Princeton University}\\
  \texttt{jacquelinehe00@gmail.com}\\
  \texttt{\{mengzhou, fellbaum, danqic\}@cs.princeton.edu}
  }

\begin{document}
\maketitle


\begin{abstract}
Pre-trained language models encode undesirable social biases, which are further exacerbated in downstream use. To this end, we propose \ours{} (a \underline{M}ethod for \underline{A}ttenuating Gender \underline{B}ias using \underline{E}ntailment \underline{L}abels), an intermediate pre-training approach for mitigating gender bias in contextualized representations. 
Key to our approach is the use of a contrastive learning objective on counterfactually augmented, gender-balanced entailment pairs from natural language inference (NLI) datasets. We also introduce an alignment regularizer that pulls identical entailment pairs along opposite gender directions closer. We extensively evaluate our approach on intrinsic and extrinsic metrics, and show that {\ours} outperforms previous task-agnostic debiasing approaches in terms of fairness. It also preserves task performance after fine-tuning on downstream tasks. Together, these findings demonstrate the suitability of NLI data as an effective means of bias mitigation, as opposed to only using unlabeled sentences in the literature. Finally, we identify that existing approaches often use evaluation settings that are insufficient or inconsistent. We make an effort to reproduce and compare previous methods, and call for unifying the evaluation settings across gender debiasing methods for better future comparison.\footnote{Our code is publicly available at \url{https://github.com/princeton-nlp/MABEL}.}

\end{abstract}


\section{Introduction}
\label{sec:introduction}

Pre-trained language models have reshaped the landscape of modern natural language processing~\cite{peters-etal-2018-deep,devlin2019bert,liu2019roberta}. As these powerful networks are optimized to learn statistical properties from large training corpora imbued with significant social biases (e.g., gender, racial), they produce encoded representations that inherit undesirable associations as a byproduct~\cite{zhao-etal-2019-gender,webster2020measuring,nadeem-etal-2021-stereoset}. More concerningly, models trained on these representations can not only propagate but also amplify discriminatory judgments in downstream applications~\cite{kurita-etal-2019-measuring}.

A multitude of recent efforts have focused on alleviating biases in language models. These can be classed into two categories (Table~\ref{tab:baseline_chart}): 1) \ti{task-specific} approaches perform bias mitigation during downstream fine-tuning, and require data to be annotated for sensitive attributes; 2) \ti{task-agnostic} approaches directly improve pre-trained representations, most commonly either by removing discriminative biases through projection~\cite{Dev_Li_Phillips_Srikumar_2020, liang-etal-2020-towards,kaneko-bollegala-2021-context}, or by performing intermediate pre-training on gender-balanced data ~\cite{webster2020measuring,ChengHYSC21,lauscher-etal-2021-sustainable-modular, guo-etal-2022-auto}, resulting in a new encoder that can transfer fairness effects downstream via standard fine-tuning.


In this work, we present {\ours}, a novel and lightweight method for attenuating gender bias. {\ours} is task-agnostic and can be framed as an intermediate pre-training approach with a contrastive learning framework. Our approach hinges on the use of entailment pairs from supervised natural language inference datasets~\cite{bowman-etal-2015-large,N18-1101}. We augment the training data by swapping gender words in both premise and hypothesis sentences and model them using a contrastive objective. We also propose an alignment regularizer, which minimizes the distance between the entailment pair and its augmented one. \ours{} optionally incorporates a masked language modeling objective, so that it can be used for token-level downstream tasks.


To the best of our knowledge, \ours{} is the first to exploit supervised sentence pairs for learning fairer contextualized representations. Supervised contrastive learning via entailment pairs is known to learn a more uniformly distributed representation space, wherein similarity measures between sentences better correspond to their semantic meanings~\cite{gao2021simcse}. Meanwhile, our proposed alignment loss, which pulls identical sentences along contrasting gender directions closer, is well-suited to learning a fairer semantic space.  


 

We systematically evaluate \ours{} on a comprehensive suite of intrinsic and extrinsic measures spanning language modeling, text classification, NLI, and coreference resolution. {\ours} performs well against existing gender debiasing efforts in terms of both fairness and downstream task performance, and it also preserves language understanding on the GLUE benchmark~\cite{wang-etal-2018-glue}. Altogether, these results demonstrate the effectiveness of harnessing NLI data for bias attenuation, and underscore \ours{}'s potential as a general-purpose fairer encoder.


Lastly, we identify two major issues in existing gender bias mitigation literature. First, many previous approaches solely quantify bias through the Sentence Encoding Association Test (SEAT)~\cite{may-etal-2019-measuring}, a metric that compares the geometric relations between sentence representations. Despite scoring well on SEAT, many debiasing methods do not show the same fairness gains across other evaluation settings. Second, previous approaches evaluate on extrinsic benchmarks in an inconsistent manner. For a fairer comparison, we either reproduce or summarize the performance of many recent methodologies on major evaluation tasks. We believe that unifying the evaluation settings lays the groundwork for more meaningful methodological comparisons in future research.

\begin{table*}[!ht]
    \centering 
    \resizebox{\textwidth}{!}{
    \begin{tabular}{@{}lcccccl@{}} \toprule
     \multirow{2}{*}{\textbf{Method}} & \textbf{Proj.} & \textbf{Con.} & \textbf{Gen.} &  \textbf{LM}  &\textbf{Fine-} &  \multirow{2}{*}{\textbf{Intermediate pre-training data}}\\
    & \textbf{based}  & \textbf{obj.} & \textbf{aug.} &\textbf{probe} & \textbf{tune} & \\ \midrule
    \textbf{Task-specific approaches} \\
    ~~\textsc{INLP} \cite{ravfogel-etal-2020-null} & \cmark         &      &           & \cmark*       & \cmark        & Wikipedia*                                  \\
      
    ~~\textsc{Con} \cite{shen-2021-contrastive}   &             & \cmark  &           &            & \cmark        & -                                  \\
    ~~\textsc{DADV}  \cite{han2021diverse}  &             &      &           &            & \cmark        & -                                  \\
    ~~\textsc{GATE}  \cite{han2021balancing}  &             &      &           &            & \cmark        & -                                  \\
     ~~\textsc{R-LACE}   \cite{ravfogel2022linear}      & \cmark         &      &           &            & \cmark        & -                                  \\
    \midrule
    \textbf{Task-agnostic approaches} \\
    ~~\textsc{CDA} \cite{webster2020measuring} &             &      & \cmark       & \cmark        & \cmark        & Wikipedia   (1M steps, 36h on 8x 16 TPU)                       \\
    ~~\textsc{Dropout} \cite{webster2020measuring}   &             &      &           & \cmark        & \cmark        & Wikipedia (100K steps, 3.5h on 8x 16 TPU)                         \\
    ~~\textsc{ADELE}  \cite{lauscher-etal-2021-sustainable-modular}  &             &      & \cmark       & \cmark        & \cmark        & Wikipedia, BookCorpus       (105M sentences)       \\
    ~~\textsc{Bias Projection} \cite{Dev_Li_Phillips_Srikumar_2020} & \cmark         &      & $\divideontimes$          & \cmark        & \cmark        & Wikisplit     (1M sentences)                    \\
    ~~\textsc{OSCaR} \cite{dev2021oscar} &  &  & $\divideontimes$  &  \cmark  & &  SNLI$^{\sharp}$ (190.1K sentences) \\
    ~~\textsc{Sent-Debias} \cite{liang-etal-2020-towards}   & \cmark         &      & \cmark       & \cmark        & \cmark        & WikiText-2, SST, Reddit, MELD, POM    \\
    ~~\textsc{Context-Debias} \cite{kaneko-bollegala-2021-context} & \cmark         &      &   $\divideontimes$        & \cmark        & \cmark        & News-commentary-v1     (87.66K sentences)            \\
    ~~\textsc{Auto-Debias}   \cite{guo-etal-2022-auto}   &             &   &        &    & \cmark        & Bias prompts generated from Wikipedia (500)\\
    ~~\textsc{\ff{}}   \cite{ChengHYSC21}   &             & \cmark  & \cmark       &  \danger    & \cmark        & WikiText-2, SST, Reddit, MELD, POM \\
    ~~\text{\ding{72}}\textsc{\ours{}} (ours)  &             & \cmark  & \cmark       & \cmark        & \cmark        & MNLI, SNLI with gender terms (134k sentences)\\ \bottomrule
    \end{tabular}}
    \caption{Properties of existing gender debiasing approaches for \textit{contextualized} representations. \textbf{Proj. based}: projection-based. \textbf{Con. obj.}: based on contrastive objectives. \textbf{Gen. aug.}: these approaches use a seed list of gender terms for counterfactual data augmentation. \textbf{LM probe} and \textbf{Fine-tune} denote that the approach can be used for language model probing or fine-tuning, respectively. $\ast$: \textsc{INLP} was originally only used for task-specific fine-tuning; \citet{meadeempirical2022} later adapted it for task-agnostic training on Wikipedia for LM probing. \danger: \textsc{\ff{}} shows poor LM probing performance in \autoref{tab:ss} as the debiasing filter is not trained with an MLM head. \textsc{\ours{}} fixes this issue by jointly training with an MLM objective. $\divideontimes$: these works use a single gender pair ``he/she'' to calculate the gender subspace. $\sharp$: \citet{dev2021oscar} fine-tunes on SNLI but does not use it for debiasing.} 

    \label{tab:baseline_chart}
\end{table*}

\section{Background}
\label{sec:background}



\subsection{Debiasing Contextualized Representations}

Debiasing attempts in NLP can be divided into two categories. In the first category, the model learns to disregard the influence of sensitive attributes in representations during fine-tuning, through projection-based~\cite{ravfogel-etal-2020-null, ravfogel2022linear}, adversarial~\cite{han2021balancing, han2021diverse} or contrastive~\cite{shen-2021-contrastive, chi-2022-conditional} downstream objectives. This approach is \textit{task-specific} as it requires fine-tuning data that is annotated for the sensitive attribute. The second type, \textit{task-agnostic} training, mitigates bias by leveraging textual information from general corpora. This can involve computing a gender subspace and eliminating it from encoded representations~\cite{Dev_Li_Phillips_Srikumar_2020, liang-etal-2020-towards, dev2021oscar, kaneko-bollegala-2021-context}, or by re-training the encoder with a higher dropout~\cite{webster2020measuring} or equalizing objectives~\cite{ChengHYSC21, guo-etal-2022-auto} to alleviate unwanted gender associations.

We summarize recent efforts of both task-specific and task-agnostic approaches in \autoref{tab:baseline_chart}. Compared to task-specific approaches that only debias for the task at hand, task-agnostic models produce fair encoded representations that can be used toward a variety of applications. \ours{} is task-agnostic, as it produces a general-purpose debiased model. 
Some recent efforts have broadened the scope of task-specific approaches. For instance,~\citet{meadeempirical2022} adapt the task-specific Iterative Nullspace Linear Projection (INLP)~\cite{ravfogel-etal-2020-null} algorithm to rely on Wikipedia data for language model probing.  While non-task-agnostic approaches can potentially be adapted to general-purpose debiasing, we primarily consider other task-agnostic approaches in this work.

\subsection{Evaluating Biases in NLP}

The recent surge of interest in fairer NLP systems has surfaced a key question: how should bias be quantified? \textit{Intrinsic} metrics directly probe the upstream language model, whether by measuring the geometry of the embedding space~\cite{caliskan2017semantics, may-etal-2019-measuring, 10.1145/3461702.3462536}, or through likelihood-scoring~\cite{kurita-etal-2019-measuring, nangia-etal-2020-crows, nadeem-etal-2021-stereoset}. \textit{Extrinsic} metrics evaluate for fairness by comparing the system's predictions across different populations on a downstream task~\cite{de-arteaga-et-al, zhao-etal-2019-gender, Dev_Li_Phillips_Srikumar_2020}. Though opaque, intrinsic metrics are fast and cheap to compute, which makes them popular among contemporary works~\cite{meadeempirical2022, qian2022perturbation}. Comparatively, though extrinsic metrics are more interpretable and reflect tangible social harms, they are often time- and compute-intensive, and so tend to be less frequently used.\footnote{As \autoref{tab:baseline_metrics_chart} in \autoref{sec:appendix_benchmarks} indicates, many previous bias mitigation approaches limit evaluation to 1 or 2 metrics.}

To date, the most popular bias metric among task-agnostic approaches is the Sentence Encoder Association Test (SEAT)~\cite{may-etal-2019-measuring}, which compares the relative distance between the encoded representations. Recent studies have cast doubt on the predictive power of these intrinsic indicators. SEAT has been found to elicit counter-intuitive results from encoders~\cite{may-etal-2019-measuring} or exhibit high variance across identical runs~\cite{aribandi-etal-2021-reliable}. ~\citet{goldfarb-tarrant-etal-2021-intrinsic} show that intrinsic metrics do not reliably correlate with extrinsic metrics, meaning that a model could score well on SEAT, but still form unfair judgements in downstream conditions. This is especially concerning as many debiasing studies~\cite{liang-etal-2020-towards, ChengHYSC21} solely report on SEAT, which is shown to be unreliable and incoherent. For these reasons, we disregard SEAT as a main intrinsic metric in this work.\footnote{For comprehensiveness, we report \ours{}'s results on SEAT in \autoref{sec:appendix_seat}.} 

Bias evaluation is critical as it is the first step towards detection and mitigation. Given that bias reflects across language in many ways, relying upon a single bias indicator is insufficient~\cite{silva-etal-2021-towards}. Therefore, we benchmark not just \ours{}, but also current task-agnostic methods against a diverse set of intrinsic and extrinsic indicators. 






\begin{figure*}[th]
\includegraphics[width=\linewidth]{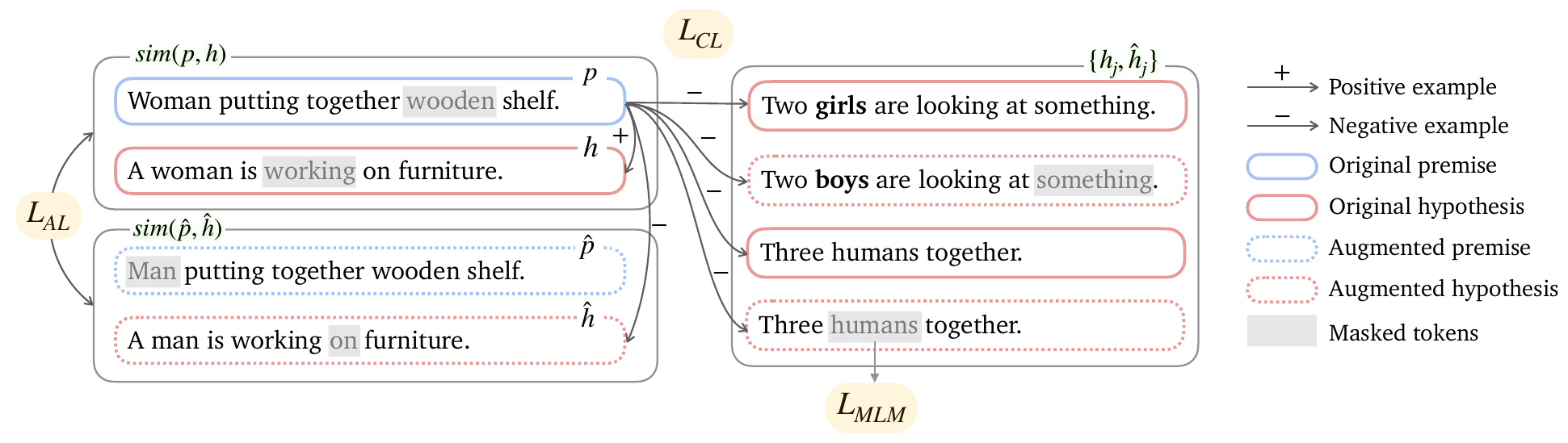}
\caption{\ours{} consists of three losses: 1) an entailment-based contrastive loss ($\mathcal{L}_{\text{CL}})$ that uses the premises's hypothesis as a positive sample and other in-batch hypotheses as negative samples; 2) an alignment loss ($\mathcal{L}_{\text{AL}})$ that minimizes the similarity difference between each original entailment pair and its gender-balanced counterpart; 3) a masked language modeling loss ($\mathcal{L}_{\text{MLM}})$ to recover $p = 15\%$ of the masked tokens.}
\label{fig:teaser}
\end{figure*}

\section{Method}
\label{sec:method}

{\ours} attenuates gender bias in pre-trained language models by leveraging entailment pairs from natural language inference (NLI) data to produce general-purpose debiased representations. To the best of our knowledge, {\ours} is the first method that exploits semantic signals from supervised sentence pairs for learning fairness.


\subsection{Training Data}

NLI data is shown to be especially effective in training discriminative and high-quality sentence representations~\cite{conneau-etal-2017-supervised,reimers2019sentence,gao2021simcse}. While previous works in fair representation learning use generic sentences from different domains~\citep{liang-etal-2020-towards, ChengHYSC21, kaneko-bollegala-2021-context}, we explore using sentence pairs with an \ti{entailment} relationship: a {hypothesis} sentence that can be inferred to be true, based on a {premise} sentence. Since gender is our area of interest, we extract all entailment pairs that contain at least one gendered term in either the premise or the hypothesis from an NLI dataset. In our experiments, we explore using two well-known NLI datasets: the Stanford Natural Language Inference (SNLI) dataset~\cite{bowman-etal-2015-large} and the Multi-Genre Natural Language Inference (MNLI) dataset~\cite{N18-1101}.



As a pre-processing step, we first conduct counterfactual data augmentation~\cite{webster2020measuring} on the entailment pairs. For any sensitive attribute term in a word sequence, we swap it for a word along the opposite bias direction, i.e., \textit{girl} to \textit{boy}, and keep the non-attribute words unchanged.\footnote{We use the same list of attribute word pairs from~\citet{bolukbasi2016man},~\citet{liang-etal-2020-towards}, and~\citet{ChengHYSC21}, which can be found in \autoref{sec:appendix_method_impl}.} This transformation is systematically applied to each sentence in every entailment pair. An example of this augmentation, with gender bias as the sensitive attribute, is shown in \autoref{fig:teaser}.



\subsection{Training Objective}

Our training objective consists of three components: a contrastive loss based on entailment pairs and their augmentations, an alignment loss, and an optional masked language modeling loss.


\paragraph{Entailment-based contrastive loss.} Training with a contrastive loss induces a more isotropic representation space, wherein the sentences' geometric positions can better align with their semantic meaning~\cite{wang2020hypersphere, gao2021simcse}. We hypothesize that this contrastive loss would be conducive to bias mitigation, as concepts with similar meanings, but along opposite gender directions, move closer under this similarity measurement. Inspired by~\citet{gao2021simcse}, we use a contrastive loss that encourages the inter-association of entailment pairs, with the goal of the encoder also learning semantically richer associations.\footnote{In this work, we only refer to the supervised SimCSE model, which leverages entailment pairs from NLI data.} 

With $p$ as the premise representation and $h$ as the hypothesis representation, let $\{(p_i, h_i)\}_{i=1}^{n}$ be the sequence of representations for $n$ original entailment pairs, and $\{(\hat p_i, \hat h_i)\}_{i=1}^{n}$ be $n$ counterfactually-augmented entailment pairs. Each entailment pair (and its corresponding augmented pair) forms a positive pair, and the other in-batch sentences constitute negative samples. With $m$ pairs and their augmentations in one training batch, the contrastive objective for an entailment pair $i$ is defined as: 
\begin{align*}
\vspace{-1em}
\mathcal{L}^{(i)}_{\textrm{CL}} =&  - \log{\frac{e^{\textrm{sim}(p_i, h_i) / \tau}}{\sum_{j=1}^{m} e^{\textrm{sim}(p_i, h_j)/ \tau}  + e^{\textrm{sim}(p_i, \hat h_j)/ \tau}}} \\
& - \log{\frac{e^{\textrm{sim}(\hat{p}_i, \hat{h}_i) / \tau}}{\sum_{j=1}^{m} e^{\textrm{sim}(\hat{p}_i, h_j)/ \tau}  + e^{\textrm{sim}(\hat{p}_i, \hat h_j)/ \tau}}},
\vspace{-1em}
\end{align*}







\noindent where $\textrm{sim}(\cdot, \cdot)$ denotes the cosine similarity function, and $\tau$ is the temperature. $\mathcal{L}_{\textrm{CL}}$ is simply the average of all the losses in a training batch. Note that when $h_i = \hat{h}_i$ (i.e., when $h_i$ does not contain any gender words and the augmentation is unchanged), we exclude $\hat{h}_i$ from the denominator to avoid $h_i$ as a positive sample and $\hat{h}_i$ as a negative sample for $p_i$, and vice versa. 

\paragraph{Alignment loss.} We want a loss that encourages the intra-association between the original entailment pairs and their augmented counterparts. Intuitively, the features from an entailment pair and its gender-balanced opposite should be taken as positive samples and be spatially close. Our alignment loss minimizes the distance between the cosine similarities of the original sentence pairs $(p_i, h_i)$ and the gender-opposite sentence pairs $(\hat p_i, \hat h_i)$:
\vspace{-0.2em}
\begin{align*}
\vspace{-2em}
   \mathcal{L_\textrm{AL}}  &= \frac{1}{m} \sum_{i=1}^m \left(\textrm{sim}(\hat p_i, \hat h_i) - \textrm{sim}(p_i, h_i)\right)^2.
   \label{eq:align_loss}
\end{align*}

We assume that a model is less biased if it assigns similar measurements to two gender-opposite pairs, meaning that it maps the same concepts along different gender directions to the same contexts.\footnote{We also explore different loss functions for alignment and report them in \autoref{sec:appendix_align}.} 





\paragraph{Masked language modeling loss.} Optionally, we can append an auxiliary masked language modeling (MLM) loss to preserve the model's language modeling capability. Following~\citet{devlin2019bert}, we randomly mask $p = 15\%$ of tokens in all sentences. By leveraging the surrounding context to predict the original terms, the encoder is incentivized to retain token-level knowledge.

In sum, our training objective is as follows:
\begin{align*}
    \scalebox{1.0}{$\mathcal{L} = (1 - \alpha) \cdot \mathcal{L_\textrm{CL}} + \alpha \cdot \mathcal{L_\textrm{AL}} + \lambda \cdot \mathcal{L_\textrm{MLM}}$},
\end{align*}
\noindent wherein the two contrastive losses are linearly interpolated by a tunable coefficient $\alpha$, and the MLM loss is tempered by the hyper-parameter $\lambda$.


\section{Evaluation Metrics}
\label{sec:evaluation}

%


\subsection{Intrinsic Metrics}
\label{subsec:intrinsic}
\paragraph{StereoSet~\cite{nadeem-etal-2021-stereoset}} queries the language model for stereotypical associations. Following~\citet{meadeempirical2022}, we consider intra-sentence examples from the gender domain. This task can be formulated as a fill-in-the-blank style problem, wherein the model is presented with an incomplete context sentence, and must choose between a stereotypical word, an anti-stereotypical word, and an irrelevant word. The Language Modeling Score (LM) is the percentage of instances in which the model chooses a valid word (either the stereotype or the anti-stereotype) over the random word; the Stereotype Score (SS) is the percentage in which the model chooses the stereotype over the anti-stereotype. The Idealized Context Association Test (ICAT) score combines the LM and SS scores into a single metric. %

\paragraph{CrowS-Pairs~\cite{nangia-etal-2020-crows}} is an intra-sentence dataset of minimal pairs, where one sentence contains a disadvantaged social group that either fulfills or violates a stereotype, and the other sentence is minimally edited to contain a contrasting advantaged group. The language model compares the masked token probability of tokens unique to each sentence. Focusing only on gender examples, we report the stereotype score (SS), the percentage in which a model assigns a higher aggregated masked token probability to a stereotypical sentence over an anti-stereotypical one.

\subsection{Extrinsic Metrics}
\label{subsec:extrinsic}
As there has been some inconsistency in the evaluation settings in the literature, we mainly consider the fine-tuning setting for extrinsic metrics and leave the discussion of the linear probing setting to \autoref{sec:appendix_probing}. 

\paragraph{Bias-in-Bios~\citep{de-Arteaga2019BiasinBios}} is a third-person biography dataset annotated by occupation and gender. We fine-tune the encoder, along with a linear classification layer, to predict an individual's profession given their biography. We report overall task accuracy and accuracy by gender, as well as two common fairness metrics~\cite{de-Arteaga2019BiasinBios, ravfogel-etal-2020-null}: 1) $GAP^{TPR}_M$, the difference in true positive rate (TPR) between male- and female-labeled instances; 2) $GAP^{TPR}_{M, y}$, the root-mean square of the TPR gap of each occupation class.

\paragraph{Bias-NLI~\cite{Dev_Li_Phillips_Srikumar_2020}} is an NLI dataset consisting of neutral sentence pairs. It is systematically constructed by populating sentence templates with a gendered word and an occupation word with a strong gender connotation (e.g., The \textit{woman} ate a bagel; The \textit{nurse} ate a bagel). Bias can be interpreted as a deviation from neutrality and is determined by three metrics: Net Neutral (NN), Fraction Neutral (FN) and Threshold:$\tau$ (T:$\tau$). A bias-free model should score a value of 1 across all 3 metrics. We fine-tune on SNLI and evaluate on Bias-NLI during inference.

\paragraph{WinoBias~\cite{zhao-etal-2018-gender}} is an intra-sentence coreference resolution task that evaluates a system's ability to correctly link a gendered pronoun to an occupation across both pro-stereotypical and anti-stereotypical contexts. Coreference can be inferred based on syntactic cues in Type 1 sentences or on more challenging semantic cues in Type 2 sentences. We first fine-tune the model on the OntoNotes 5.0 dataset~\citep{hovy-etal-2006-ontonotes} before evaluating on the WinoBias benchmark. We report the average F1-scores for pro-stereotypical and anti-stereotypical instances, and the true positive rate difference in average F1-scores, across Type 1 and Type 2 examples.  

\vspace{-0.5em}
\subsection{Language Understanding}
To evaluate whether language models still preserve general linguistic understanding after bias attenuation, we fine-tune them on seven classification tasks and one regression task from the General Language Understanding Evaluation (GLUE) benchmark~\cite{wang-etal-2018-glue}.\footnote{We also evaluate transfer performance on the SentEval tasks~\cite{conneau-etal-2017-supervised} in \autoref{sec:appendix_senteval}.}

\section{Experiments}
\label{sec:results}

\subsection{Baselines \& Implementation Details}

We choose Sent-Debias\footnote{\url{https://github.com/pliang279/sent_debias}}~\cite{liang-etal-2020-towards}, Context-Debias\footnote{\url{https://github.com/kanekomasahiro/context-debias}}~\cite{kaneko-bollegala-2021-context}, and \ff{}\footnote{As there is no code released, we use our own implementation without an auxiliary regularization term.}~\cite{ChengHYSC21} as our primary baselines. By introducing a general-purpose method for producing debiased representations, these three approaches are most similar in spirit to \ours{}. We consider \ff{} to be especially relevant as it is also a task-agnostic, contrastive learning approach. Compared to \ff{}, \ours{} leverages NLI data, and also applies entailment-based and MLM losses to ensure that sentence- and token-level knowledge is preserved. 

We evaluate the three aforementioned task-agnostic baselines across all bias benchmarks and also compare against other approaches in \autoref{tab:baseline_chart} by reporting the recorded numbers from their original work. Unless otherwise specified, all models, including \ours{}, default to \texttt{bert-base-uncased}~\cite{devlin2019bert} as the backbone encoder. In the default setting, $\lambda = 0.1$ and $\alpha = 0.05$. Implementation details on \ours{} and the task-agnostic baselines can be found in \autoref{sec:appendix_method_impl} and \autoref{sec:appendix_baseline_impl}, respectively. For our own implementations, we report the average across 3 runs.\footnote{Standard deviations can be found in \autoref{sec:appendix_std}.} 

\subsection{Results: Intrinsic Metrics}
\begin{table}[t]
\centering
\setlength{\tabcolsep}{5pt} 
\resizebox{0.49\textwidth}{!}{
\begin{tabular}{lcccc} 

\toprule
& \multicolumn{3}{c}{\tf{StereoSet}} & \tf{CrowS-Pairs} \\ 
\cmidrule(lr){2-4} \cmidrule(lr){5-5}   
\textbf{Model}   & \textbf{LM $\uparrow$} & \textbf{SS} $\diamond$ & \textbf{ICAT $\uparrow$}  &\textbf{SS} $\diamond$  \\

\midrule
\textsc{BERT}   & {84.17}     & 60.28        & 66.86      & 57.25 \ua{7.25}  \\ \midrule
\textsc{BERT+Dropout}$^{\star}$     & 83.04     & 60.66        & 65.34 & 55.34 \ua{5.34} \\
\textsc{BERT+CDA}$^{\star}$          & 83.08     & 59.61        & 67.11         & 56.11  \ua{6.11} \\
\textsc{INLP}$^{\star}$ & 80.63     & 57.25        & 68.94 & 51.15   \ua{1.15}\\
\textsc{Sent-Debias}$^{\star}$ & 84.20      & 59.37        & 68.42 & 52.29  \ua{2.29} \\
\textsc{Context-Debias}              & \tf{85.42}     & 59.35       & 69.45  & 58.01  \ua{8.01} \\
\textsc{Auto-Debias}$^{\ddagger}$ &   -  &   -     & - & 54.92   \ua{4.92}\\
\textsc{\ff{}} & 44.85 & \tf{50.93} & 44.01 & 49.03 \ua{0.97} \\
 \midrule
\textsc{\ours{}} (ours)               & 84.80              & 56.92    & \tf{73.07}               & \tf{50.76}          \ua{\tf{0.76}}               \\ 
\bottomrule
\end{tabular}}
\caption{Results on StereoSet and CrowS-Pairs (standard deviations are in \autoref{apptab:ss}). ${\star}$: the results are reported in \citet{meadeempirical2022}; ${\ddagger}$: the results are reported in \citet{guo-etal-2022-auto}. $\diamond$: the closer to 50, the better. LM: language modeling score, SS: Steoreotype score, ICAT: combined score, defined as $\text{LM} \cdot (\min(\text{SS}, 100-\text{SS})) / 50$.
}
\label{tab:ss}
\end{table}

As \autoref{tab:ss} shows, \ours{} strikes a good balance between language modeling and fairness with the highest ICAT score. Compared to BERT, \ours{} retains and even exhibits an average modest improvement (from 84.17 to 84.80) in language modeling. \ours{} also performs the best on CrowS-Pairs, with an average metric score of 50.76. 

While \ours{} does not have the best SS value for StereoSet, we must caution that this score should not be considered in isolation. For example, although \ff{} shows a better stereotype score, its language modeling ability (as the LM score shows) is significantly deteriorated and lags behind other approaches. This is akin to an entirely random model that obtains a perfect SS of 50 as it does not contain bias, but would also have a low LM score as it lacks linguistic knowledge.  

\subsection{Results: Extrinsic Metrics}

\begin{table}[!t]
\centering
\resizebox{0.50\textwidth}{!}{
\begin{tabular}{lcccccl} \toprule
      &  \tf{Acc.}  & \tf{Acc.} & \tf{Acc.} & \tf{TPR} & \tf{TPR} \\ 
\tf{Model}   & \small{\tf{(All)} $\uparrow$} & \small{\tf{(M)} $\uparrow$} & \small{\tf{(F)} $\uparrow$} & \small{\tf{GAP} $\downarrow$} & \small{\tf{RMS}} $\downarrow$ \\\midrule
BERT   & 84.14        & 84.69   & 83.50    & 1.189  & 0.144   \\ 
\midrule
\textsc{INLP}$^{\flat\dagger}$  & 70.50 & - & - & - & \tf{0.067} \\
\textsc{Con}$^{\star\dagger}$ & 81.69 & - & - & - & 0.168 \\
\textsc{DADV}$^{\sharp\dagger}$ & 81.10 & - & - & - & 0.126 \\
\textsc{GATE}$^{\sharp\dagger}$ & 80.50 & - & - & - & 0.111 \\ 
\textsc{R-LACE}$^{\flat\dagger}$ & \tf{85.04} & - & - & - & 0.115    \\
\textsc{Sent-Debias}  & 83.56       & 84.10   & 82.92   & 1.180  & 0.144   \\
\textsc{Context-Debias}   & 83.67       & 84.08    & 83.18   & 0.931  & 0.137   \\
\textsc{\ff{}} &  83.18        & 83.52    & 82.78   & 0.746   & 0.142  \\ \midrule
\textsc{\ours{}} (ours) &  84.85     & \tf{84.92} & \tf{84.34}   & \tf{0.599} & 0.132\\ 
\bottomrule
\end{tabular}}
\caption{Results on fine-tuning with the Bias in Bios dataset. $\star$: the results are reported in \citet{shen-2021-contrastive}; $\sharp$: the results are reported in \citet{han2021balancing}; $\flat$: the results are reported in \citet{ravfogel2022linear}; $\dagger$: the approaches depend on gender annotations.} 
\label{tab:occ-cls}
\end{table}

\begin{table}[t]
\centering
\resizebox{0.50\textwidth}{!}{%
\begin{tabular}{lcccc} \toprule
      \tf{Model} & \tf{TN} $\uparrow$ & \tf{FN} $\uparrow$ & \tf{T:0.5} $\uparrow$ & \tf{T:0.7}  $\uparrow$ \\ \midrule
\textsc{BERT}                 & 0.799 & 0.879 & 0.874 & 0.798\\
\midrule
\textsc{ADELE}$^\star\ddagger$ & 0.557 & 0.504 & - & -\\ 
\textsc{Sent-Debias}               & 0.793 & 0.911 & 0.897 & 0.788 \\
\textsc{Context-Debias}             & 0.858 & 0.906 & 0.902 & 0.857 \\
\textsc{Context-Debias}$^\star\ddagger$       &  0.878 & 0.968 & - & 0.893   \\
\textsc{\ff{}}               & 0.829 & 0.883 & 0.846 & 0.845 \\ \midrule
\textsc{\ours{}} (ours) & \textbf{0.900} & \textbf{0.977} & \textbf{0.974} & \textbf{0.935}\\\bottomrule
\end{tabular}}
\caption{Results on Bias-NLI. We fine-tune the models on SNLI and then evaluate on Bias-NLI. $\star$: results are reported from original papers; $\ddagger$: the models are fine-tuned on MNLI.}
\label{tab:nli}
\end{table}


\paragraph{Bias-in-Bios.}
As \autoref{tab:occ-cls} indicates, \ours{} exhibits the highest overall and individual accuracies, as well as the smallest TPR-GAP when compared against the task-agnostic baselines and BERT. 

Still, \ours{} and the other task-agnostic models are close in performance to BERT on Bias-in-Bios, which suggests that the fine-tuning process can significantly change a pre-trained model's representational structure to suit a specific downstream task. Furthermore, \citet{kaneko2022debiasing} finds that debiased language models can still re-learn social biases after standard fine-tuning on downstream tasks, which may explain why the task-agnostic methods, which operate upstream, cope worse with this particular manifestation of gender bias than on others. Our results also show that task-specific interventions fare better fairness-wise on this task. Methods such as INLP~\cite{ravfogel-etal-2020-null}, GATE~\cite{han2021balancing}, and R-LACE~\cite{ravfogel2022linear} exhibit better TPR RMS scores, although sometimes at the expense of task accuracy. As these methods operate directly on the downstream task, they may have a stronger influence on the final prediction \cite{jin-etal-2021-transferability}.

\paragraph{Bias-NLI.} We next move to Bias-NLI, where \autoref{tab:nli} indicates that \ours{} outperforms BERT and other baselines across all metrics. Unlike in Bias-in-Bios, the results have a greater spread, and \ours{}'s comparative advantage becomes clear. The FN score denotes that, on average, \ours{} correctly predicts neutral $97.7\%$ of the time. \ours{} is also more confident in predicting the correct answer, surpassing the 0.7 threshold $93.5\%$ of the time. Other approaches, such as Sent-Debias and \ff{}, do not show as clear-cut of an improvement over BERT, despite scoring well on other bmetrics such as SEAT.

As natural language inference requires robust semantic reasoning capabilities to deduce the correct answer, it is a more challenging problem than classification. Therefore, for this task, the models' initialization weights---which store the linguistic knowledge acquired in pre-training---may play a larger impact on the final task accuracy than in Bias-in-Bios.

\paragraph{WinoBias.} On this token-level extrinsic task (\autoref{tab:coref}), \ours{}, and the other bias mitigation baselines, achieve very similar average F1-scores on OntoNotes. However, performance on WinoBias becomes variegated. \ours{} shows the best task improvement on anti-stereotypical tasks, with an average 7.25\% and 10.58\% increase compared to BERT on Type 1 and Type 2 sentences, respectively. The strong performance on anti-stereotypical examples implies that \ours{} can effectively weaken the stereotypical token-level associations between occupation and gender. Though \ours{} exhibits a marginally lower F1-score on Type 1 pro-stereotypical examples (an 1.64\% decrease on average compared to the best-performing model, BERT), it has the highest F1-scores across all other categories. Furthermore, it has the best reduction in fairness, with the smallest average TPR-1 and TPR-2 by a clear margin (respectively, 23.73 and 3.41, compared to the next-best average TPR scores at 26.14 and 9.57).


\begin{table*}[!t]
\centering
\resizebox{0.82\textwidth}{!}{
\begin{tabular}{lccccccc} \toprule
 \tf{Model} & \tf{OntoNotes} $\uparrow$ & \tf{1A} $\uparrow$ & \tf{1P} $\uparrow$ & \tf{2A} $\uparrow$ & \tf{2P} $\uparrow$ & \tf{TPR-1} $\downarrow$ & \tf{TPR-2} $\downarrow$  \\  \midrule
\textsc{BERT}       &         \tf{73.53} &  53.96 & \tf{86.57} & 82.20 & 94.67       & 32.79              & 12.48              \\ 
 \midrule
 \textsc{Sent-Debias}  &     72.36       & 54.11     & 85.09         & 83.29                     & 94.73               & 30.98            & 11.44            \\
\textsc{Context-Debias}               & 73.16      & 59.40     & 85.54      & 83.63        & 93.20              & 26.14            & 9.57              \\
\textsc{\ff{}}              & 71.79       & 53.24     &   85.77      & 77.37                 & 91.40               & 32.43                  & 14.03                  \\ \midrule


\textsc{\ours{}} (ours) & 73.48 & \tf{61.21} & 84.93 & \tf{92.78} & \tf{96.20} & \tf{23.73} & \tf{3.41} \\\bottomrule
\end{tabular}}
\caption{Average F1-scores OntoNotes and WinoBias, and TPR scores across Winobias categories. 1 = Type 1; 2 = Type 2. A=anti-stereotypical; P=pro-stereotypical.}
\label{tab:coref}
\end{table*}






\begin{table*}[t]
\centering
\resizebox{1.0\textwidth}{!}{
\begin{tabular}{lccccccccc} \toprule
  & \tf{CoLA} $\uparrow$ & \tf{SST-2} $\uparrow$ & \tf{MRPC} $\uparrow$     & \tf{QQP} $\uparrow$      & \tf{MNLI} $\uparrow$ & \tf{QNLI} $\uparrow$ & \tf{RTE} $\uparrow$ & \tf{STS-B} $\uparrow$ & \tf{Avg.} $\uparrow$    \\ 
\tf{Model} & (mcc.) & (acc.) & (f1/acc.) & (acc./f1) & (acc.) & (acc.) & (acc.) & (pears./spear.)  \\ \midrule
\textsc{BERT}   & 56.5 & 92.3  & \tf{89.5}/\tf{85.3} & 90.7/87.5 & 84.3 & \textbf{92.2} & 65.0 & 88.4/88.2 & 81.8 \\ 
{\bertft}  & \bf{58.6} & \bf{93.6} & 89.4/85.1 & 90.4/86.8 & 83.3 & 89.0 & \bf{69.0} & 88.3/87.9 & \tf{82.1} \\\midrule
\textsc{Sent-Debias} & 50.5 & 89.1  & 87.5/81.6 & 87.5/90.7 & 83.9 & 91.4 & 63.2 & 88.1/87.9 & 79.4 \\
\textsc{Context-Debias}  & 55.2 & 92.0  & 85.1/77.5 & 90.7/87.4 & 84.6 & 89.9 & 57.0 & 88.4/88.1 & 79.4 \\
\textsc{\ff{}} & 55.5 & 92.4  & 87.5/80.6 & \bf{91.2}/\bf{88.1} & \bf{84.8} & 91.3 & 63.2 & 88.4/88.1 & 80.9\\ \midrule
\textsc{\ours{}} (ours) & 57.8 & 92.2 & \tf{89.5}/85.0 & \bf{91.2}/\bf{88.1} & 84.5 & 91.6 & 64.3 & \bf{89.6}/\bf{89.2} & 82.0
\\\bottomrule
\end{tabular}}
\caption{Fine-tuning results on the GLUE benchmark. BERT-NLI denotes that we fine-tune pre-trained BERT on NLI data first before fine-tuning on a GLUE task. For the average, we report the Matthew's correlation coefficient for CoLA, the Spearman's rank correlation coefficient for STS-B, and the accuracy for all other tasks.}
\label{tab:glue}
\end{table*}


\begin{figure*}[t]
\centering
\includegraphics[scale=0.36]{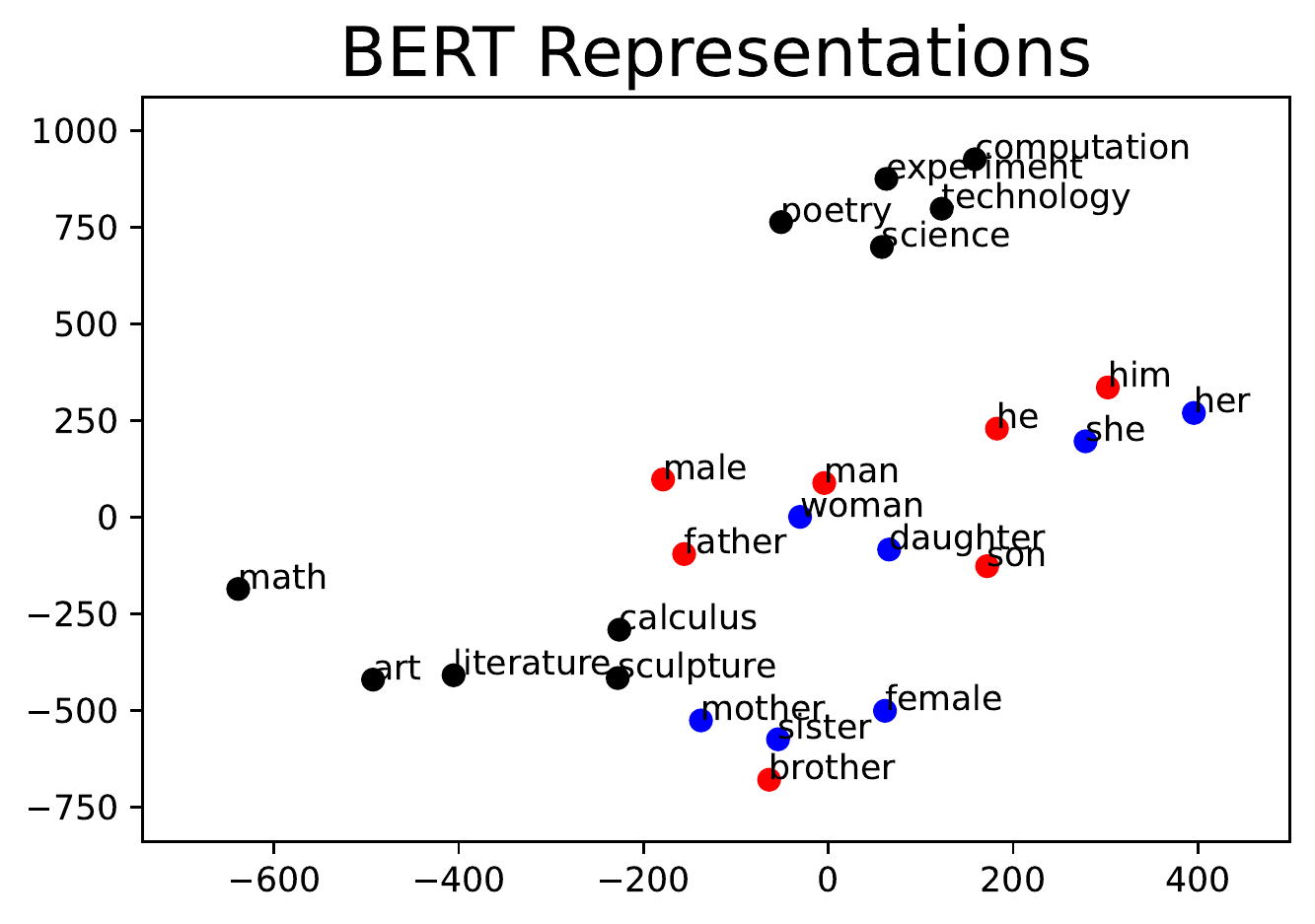}
\includegraphics[scale=0.36]{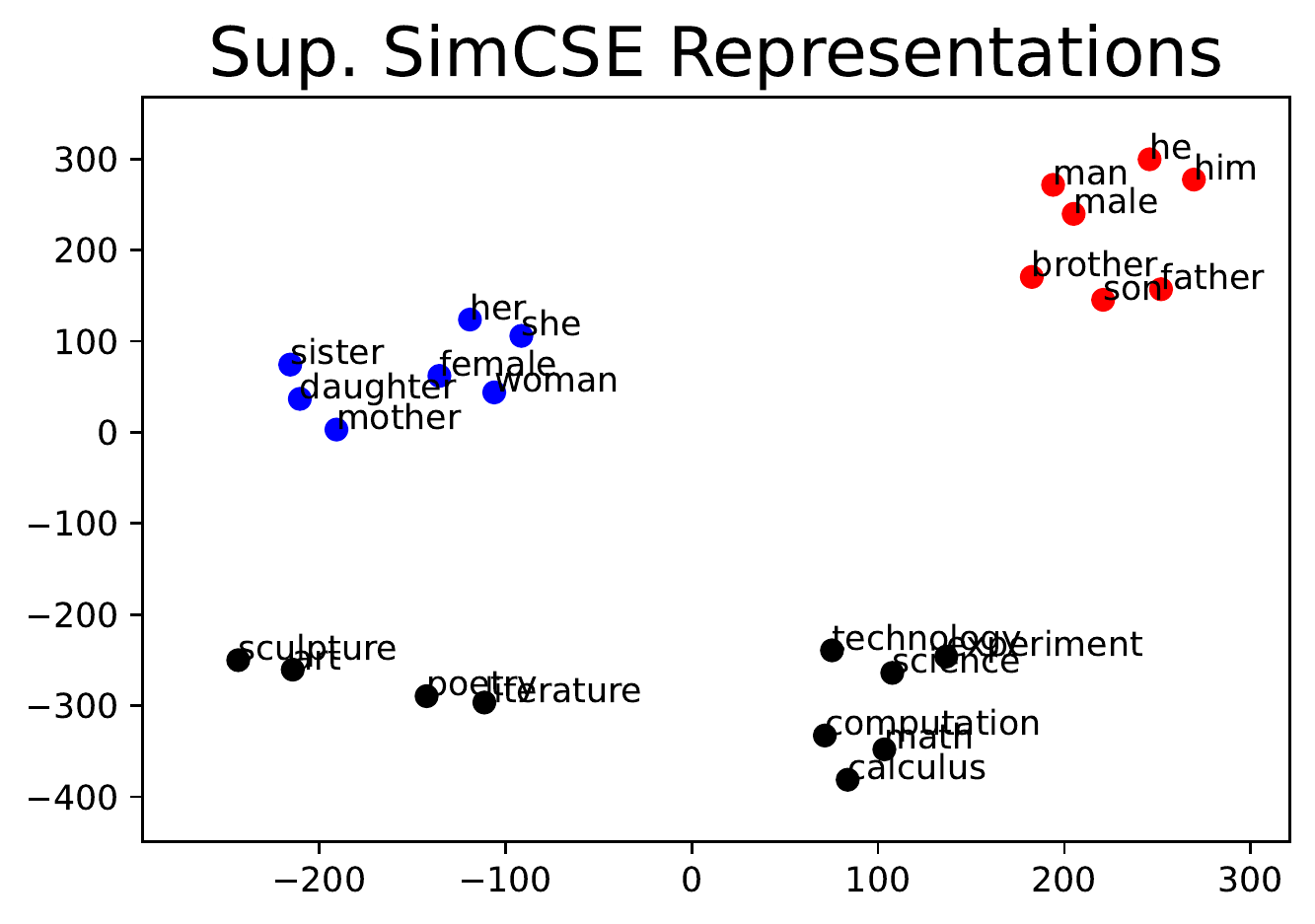}
\includegraphics[scale=0.36]{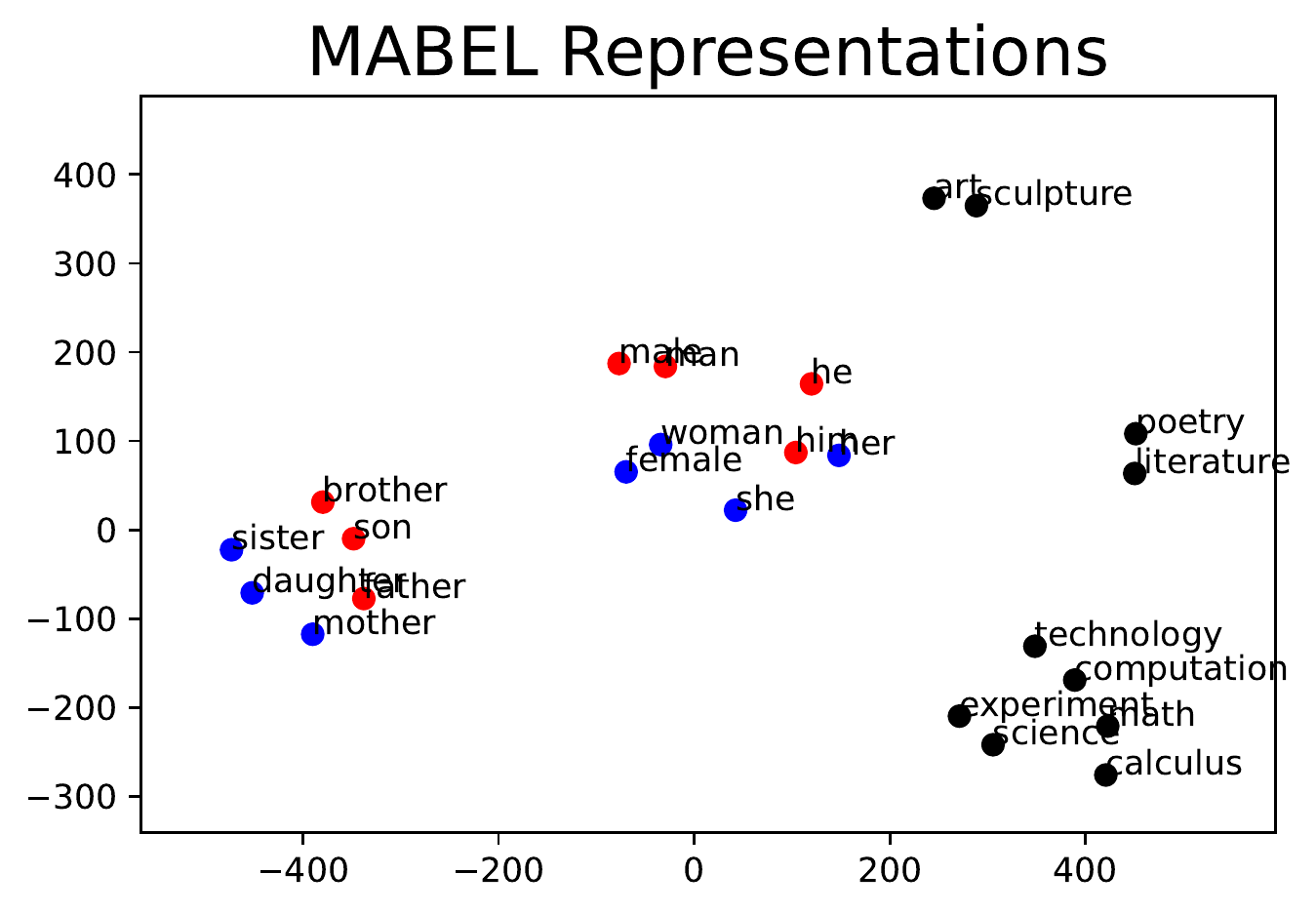}
\caption{t-SNE plots of sentence representations encoded with BERT, sup. SimCSE, and \ours{}. Male-aligned terms (man, male, he, brother, son, father) are in red, female-aligned terms (woman, female, she, her, sister, daughter, mother) are in blue. Neutral terms (e.g., math, art, calculus, poetry, science) are in black.}
\label{fig:tsne}
\end{figure*}

\subsection{Results: Language Understanding}
As the GLUE benchmark results indicate (\autoref{tab:glue}), \ours{} preserves semantic knowledge across downstream tasks. On average, \ours{} performs marginally better than BERT (82.0\% vs. 81.8\%), but not as well as BERT fine-tuned beforehand on the NLI task with MNLI and SNLI data (BERT-NLI), at 82.0\% vs. 82.1\%. Other bias mitigation baselines lag behind BERT, but the overall semantic deterioration remains minimal.



\section{Analysis}
\label{sec:analysis}

\subsection{Qualitative Comparison}
\label{app:visualization}

We perform a small qualitative study by visualizing the t-SNE \cite{JMLR:v9:vandermaaten08a} plots of sentence representations from BERT, supervised SimCSE~\cite{gao2021simcse}, and \ours{}. Following \citet{liang-etal-2020-towards, ChengHYSC21}, we plot averaged sentence representations of a gendered or neutral concept across different contexts (sentence templates). We re-use the list of gender words, and neutral words with strong gender connotations, from \citet{caliskan2017semantics}.

From \autoref{fig:tsne}, in BERT, certain concepts from technical fields such as `technology' or `science' are spatially closer to `man,' whereas concepts from the humanities such as `art' or `literature' are closer to `woman.' After debiasing with \ours{}, we observe that the gendered tokens (e.g., `man' and `woman,' or `girl' and `boy') have shifted closer in the embedding space, and away from the neutral words. While SimCSE shows a similar trend in pulling gendered words away from the neutral words, it also separates the masculine and feminine terms into two distinct clusters. This is undesirable behavior, as it suggests that identical concepts along opposite gender directions are now further apart in latent space, and are have become more differentiated in the same contexts.

\subsection{Ablations}


We perform extensive ablations to show that every component of \ours{} benefits the overall system.  We use StereoSet, CrowS-Pairs, and Bias-NLI as representative tasks.

\paragraph{Comparing other supervised pairs.} Since leveraging entailment examples as positive pairs is conducive to high-quality representation learning~\cite{gao2021simcse}, we believe that this construction is particularly suitable for semantic retention. To justify our choice, we further train on neutral pairs and contradiction pairs from the SNLI dataset. We also consider paraphrase pairs from the Quora Question Pairs (QQP) dataset~\cite{wang2017bilateral} and the Para-NMT dataset~\cite{wieting-gimpel-2018-paranmt}. Finally, we try individual unlabeled sentences from the same multi-domain corpora used by ~\citet{liang-etal-2020-towards} and~\citet{ChengHYSC21}. In this setting, standard dropout is applied: positive pairs are constructed by encoding the same sentence twice with different masks, resulting in two minimally different embeddings~\cite{gao2021simcse}. 

\begin{table}[!t]
\setlength{\tabcolsep}{3pt}
\resizebox{\linewidth}{!}{\begin{tabular}{lccccccc} \toprule
& \multicolumn{3}{c}{\tf{StereoSet}} & \tf{CSP} & \multicolumn{3}{c}{\tf{Bias-NLI}} \\
\cmidrule(lr){2-4} \cmidrule(lr){5-5} \cmidrule(lr){6-8}
      & {LM}$\uparrow$ & {SS}$\diamond$  & {ICAT}$\uparrow$  & {SS}$\diamond$   & {NN}$\uparrow$   & {FN}$\uparrow$  & {TN:0.5}$\uparrow$  \\ \midrule
\textsc{Default} & 84.5                 & 56.2                  & 74.0                    & \tf{50.8}                  & 0.917                  & \tf{0.983}                  & \tf{0.983}                      \\
\textsc{SNLI Ent.}    & 84.1                  & 58.9                  & 69.1                    & 51.5                  & 0.885                  & 0.973                  & 0.972                      \\
\textsc{MNLI Ent.}    & \tf{85.8}                  & \tf{55.7}                  & \tf{76.1}                     & 53.8                  & 0.915                  & 0.927                  & 0.971                      \\ \midrule
\textsc{SNLI Neu.} & 82.8                  & 58.9                  & 68.3                    & 55.0                  & \tf{0.935}                  & 0.945                  & 0.945                      \\
\textsc{SNLI Con.} & 76.9                  & 58.0                  & 64.6                    & 56.5                  & 0.710                   & 0.723                  & 0.722                      \\ \midrule
\textsc{QQP}       & 76.9                  & 57.9                  & 64.6                    & 53.1                  & 0.917                  & 0.938                  & 0.938                      \\
\textsc{Para-NMT}  & 79.3                  & 57.8                  & 67.0                    & 53.4                  & 0.756                  & 0.783                  & 0.782                      \\
\textsc{Dropout}   & 78.4                  & 57.3                  & 67.0                    & 52.7                  & 0.780                   & 0.809                  & 0.807                      \\ \bottomrule
\end{tabular}}
\caption{Data ablation results for \ours{}. Positive pair constructions include entailment (Ent.), neutral (Neu.) and contradictory (Con.) pairs, paraphrastic examples from QQP and Para-NMT, and general sentences from the corpora used by \citet{liang-etal-2020-towards}. Default: SNLI+MNLI entailment data. Dropout: the same sentence is passed through the encoder twice with standard dropout.  CSP: CrowS-Pairs. $\diamond$: the closer to 50, the better.}
\label{tab:aba_datasets}
\end{table}

From \autoref{tab:aba_datasets}, entailment pairs are a critical data choice for preserving language modeling ability, and result in the highest ICAT scores. Interestingly, the SS is consistent across the board. The MNLI dataset produces the best LM and SS scores, likely as it is a semantically richer dataset with sentences harvested across diverse genres. In contrast, SNLI consists of short, artificial sentences harvested from image captions. The exposure to MNLI's diverse vocabulary set may have helped \ours{} learn better language modeling and greater fairness across a broad range of semantic contexts. The stereotype scores from StereoSet and CrowS-Pairs do not correlate well; for instance, \ours{} trained on SNLI entailment pairs shows the worst stereotype score on StereoSet, but among the best on CrowS-Pairs. With only 266 examples, CrowS-Pairs is significantly smaller than StereoSet (which has 2313 examples), and tends to be a more equivocal metric. Entailment pairs, and neutral pairs to a lesser extent, demonstrate the best language retention on the Bias-NLI metric, although the QQP dataset also performs well. One possible explanation is that the QQP paraphrases hold a similar, albeit weaker, directional relationship to NLI pairs.    

\paragraph{Disentangling objectives.} Results of \ours{} trained with ablated losses are in \autoref{tab:aba_objective}. Without the MLM objective, the LM score collapses along with the ICAT score. Though the SS score approaches 50, it seems to be more indicative of randomness than model fairness; the off-the-shelf LM head is no longer compatible with the trained encoder. With the contrastive loss omitted, \ours{}'s performance on Bias-NLI drops from the 0.9 range to the 0.8 range, showing that it is key to preserving sentence-level knowledge. Removing the alignment loss also leads to a similar decrease in performance on Bias-NLI. As this particular objective does not directly optimize semantic understanding, we attribute this drop to a reduction in fairness knowledge.

\begin{table}[!t]
\setlength{\tabcolsep}{3pt}
\resizebox{1.0\linewidth}{!}{
\begin{tabular}{@{}lccccccc@{}} \toprule
& \multicolumn{3}{c}{\tf{StereoSet}} & \tf{CSP} & \multicolumn{3}{c}{\tf{Bias-NLI}} \\
\cmidrule(lr){2-4} \cmidrule(lr){5-5} \cmidrule(lr){6-8}
      & {LM}$\uparrow$ & {SS}$\diamond$  & {ICAT}$\uparrow$  & {SS}$\diamond$   & {NN}$\uparrow$   & {FN}$\uparrow$  & {TN:0.5}$\uparrow$  \\ \midrule
\textsc{MABEL} & 84.6 & 56.2 & \tf{74.0} & \tf{50.8} & 0.917 & \tf{0.983} & \tf{0.982}  \\ 
~~$- {\mathcal{L}_{\textsc{MLM}}}$  & 55.8 & \tf{51.1} & 54.6 & 44.3 & \tf{0.970}  & 0.976 & 0.976  \\
~~${- \mathcal{L}_{\textsc{CL}}}$   & 84.9 & 57.2 & 72.6 & 54.6  & 0.858 & 0.884 & 0.883  \\
~~$-{\mathcal{L}_{\textsc{AL}}}$   & \tf{85.0} & 57.3 & 72.6 & 54.2  & 0.878 & 0.890 & 0.889  \\ \bottomrule
\end{tabular}}
\caption{Objective ablation results for \ours{}. CSP: CrowS-Pairs. $\diamond$: the closer to 50, the better.
\label{tab:aba_objective}
}
\end{table}


\paragraph{Impact of batch size.} \autoref{tab:app_tuning} shows the effect of batch size on StereoSet performance. Encouragingly, although contrastive representation learning typically benefits from large batch sizes~\cite{chen2020simple}, an aggregated batch size of 128 already works well. \ours{} is very lightweight and trains in less than 8 hours on a single GPU.
\begin{table}[!t]
\centering
\resizebox{0.4\textwidth}{!}{
\begin{tabular}{cccc}
\toprule
 \bf{Batch Size} &  \bf{LM} $\uparrow$ & \bf{SS} $\diamond$  & \bf{ICAT} $\uparrow$ \\ \midrule
$64$  & 82.43 & 56.42 & 71.85\\
$128$  & 84.55 & \bf{56.25} & \bf{73.98}\\
$256$  & \tf{84.62} & 57.46 & 72.00\\ \bottomrule
\end{tabular}}
\caption{StereoSet results on different cumulative batch sizes. $\diamond$: the closer to 50, the better.} 
\label{tab:app_tuning}
\end{table}

\section{Conclusion}
\label{sec:conclusion}


We propose \ours{}, a simple bias mitigation technique that harnesses supervised signals from entailment pairs in NLI data to create informative and fair contextualized representations. We compare \ours{} and other recent task-agnostic debiasing baselines across a wide range of intrinsic and extrinsic bias metrics, wherein \ours{} demonstrates a better performance-fairness tradeoff. Its capacity for language understanding is also minimally impacted, rendering it suitable for general-purpose use. Systematic ablations show that both the choice of data and individual objectives are integral to \ours{}'s good performance. Our contribution is complementary to the bias transfer hypothesis \cite{jin-etal-2021-transferability}, which suggests that upstream bias mitigation effects are transferable to downstream settings. We hope that \ours{} adds a new perspective toward creating fairer language models. 





\clearpage
\section*{Acknowledgements}



We thank Tianyu Gao, Sadhika Malladi, Howard Yen, and the other members of the Princeton NLP group for their helpful discussions and support. We are also grateful to the anonymous reviewers for their valuable feedback. This research is partially supported by the Peter and Rosalind Friedland Endowed Senior Thesis Fund from the Princeton School of Engineering \& Applied Sciences.

\section*{Limitations}
\label{sec:limitations}

Following prior bias mitigation work \cite{ ChengHYSC21, liang-etal-2020-towards}, our framework relies on a curated list of gender word pairs for counterfactual data augmentation. While we believe that our general list is broad enough to cover the majority of gendered terms in a dataset, this lexicon is nevertheless non-exhaustive and cannot completely remove all bias directions \cite{ethayarajh-etal-2019-understanding}. One possible improvement would be to use automatic perturbation augmentation on the entailment pairs \cite{qian2022perturbation} (concurrent work), a more expansive technique that counterfactually augments data along multiple demographic axes. 

Another consideration is that we primarily juxtapose against task-agnostic approaches in our work, even though some task-specific procedures, specifically R-LACE \cite{ravfogel2022linear} and INLP \cite{ravfogel-etal-2020-null}, show excellent gains in occupation classification, an extrinsic task. Recently, \citet{meadeempirical2022} has successfully adapted INLP to a task-agnostic setting by mining on an unlabeled corpus. We believe that other task-specific methods can be similarly adapted to train task-agnostic encoders, though we leave this comparison to future work. In light of recent findings that bias can re-enter the model during any stage of the training pipeline \cite{jin-etal-2021-transferability, kaneko2022debiasing}, one interesting direction would be to pair \ours{}, which is task-agnostic, with task-specific procedures. Essentially, by debiasing at both ends---first upstream in the encoder, then downstream in the classifier---\ours{} could potentially achieve a greater reduction in bias across some of the extrinsic benchmarks.

Although \ours{} shows exciting performance across an extensive range of evaluation settings, these results should not be construed as a complete erasure of bias. For one, our two main intrinsic metrics, StereoSet and CrowS-Pairs, are skewed towards North American social biases and only reflect positive predictive power. They can detect the presence, not the \textit{absence} of bias \cite{meadeempirical2022}. \citet{aribandi-etal-2021-reliable} furthers that these likelihood-based diagnostics can vary wildly across identical model checkpoints trained on different random seeds. \citet{blodgett-etal-2021-stereotyping} points to the unreliability of several benchmarks we use, including StereoSet, CrowS-Pairs, and WinoBias, which inadequately articulate their assumptions of stereotypical behaviors. Additionally, \ours{}'s gains in fairness are not universally strong---it handles some operationalizations of gender bias more effectively than others. One reason for this inconsistency is that bias metrics have been found to correlate poorly; desirable performance on one bias indicator does not necessarily translate to equivalently significant gains on other evaluation tasks~\cite{goldfarb-tarrant-etal-2021-intrinsic, orgad2022how}. The lack of clarity and agreement in existing evaluation frameworks is a fundamental challenge in this field.




\section*{Ethics Statement}

There are several ethical points of consideration to this work. As our contribution is entirely methodological, we rely upon an existing range of well-known datasets and evaluation tasks that assume a binary conceptualization of gender. In particular, the over-simplification of gender identity as a dichotomy, not as a spectrum, means that \ours{} does not adequately address the full range of stereotypical biases expressed in real life. We fully acknowledge and support the development of more inclusive methodological tools, datasets, and evaluation mechanisms. 

Furthermore, we restrict the definitonal scope of bias in this work to allocational and representational bias \cite{barocas2017problem}. \textit{Allocational bias} is the phenomenon in which models perform systematically better for some social groups over others, e.g., a coreference resolution system that successfully identifies male coreferents at a higher rate over female ones. \textit{Representational} bias denotes the spurious associations between social groups and certain words or concepts. An example would be the unintentional linkage of genders with particular occupations, as captured by contextualized word representations. 

We neglect other critical types of biases under this framework, in particular intersectional biases. As per \citet{subramanian-etal-2021-evaluating}, most existing debiasing techniques only consider sensitive attributes, e.g., race or gender, in isolation. However, a truly fair model does not and cannot operate in a vacuum, and should be able to handle a complex combination of various biases at once.

Another consideration is that \ours{} is entirely English-centric. This assumption is symptomatic of a larger problem, as most gender bias studies are situated in high-resource languages. Given that conceptualizations of gender and language are a function of societal and cultural norms, it is imperative that the tools we create can generalize beyond an English context. For instance, some languages such as Spanish or German contain grammatical gender, meaning that nouns or adjectives can have masculine or feminine forms. The need to account for both linguistic gender and social gender significantly complicates the matter of bias detection and elimination.

For these reasons, practitioners should exercise great caution when applying \ours{} to real-world use cases. At its present state, \ours{} should not be viewed as a one-size-fits-all solution to gender bias in NLP, but moreso as a preliminary effort to illuminate and attenuate aspects of a crucial, elusive, and multi-faceted problem.


\bibliography{ref}

\begin{thebibliography}{69}
\expandafter\ifx\csname natexlab\endcsname\relax\def\natexlab#1{#1}\fi

\bibitem[{Aribandi et~al.(2021)Aribandi, Tay, and
  Metzler}]{aribandi-etal-2021-reliable}
Vamsi Aribandi, Yi~Tay, and Donald Metzler. 2021.
\newblock \href {https://doi.org/10.18653/v1/2021.findings-acl.155} {How
  reliable are model diagnostics?}
\newblock In \emph{Findings of the Association for Computational Linguistics:
  ACL-IJCNLP 2021}, pages 1778--1785, Online. Association for Computational
  Linguistics.

\bibitem[{Barocas et~al.(2017)Barocas, Crawford, Shapiro, and
  Wallach}]{barocas2017problem}
Solon Barocas, Kate Crawford, Aaron Shapiro, and Hanna Wallach. 2017.
\newblock The problem with bias: from allocative to representational harms in
  machine learning.
\newblock In \emph{SIGCIS Conference}, Online.

\bibitem[{Bartl et~al.(2020)Bartl, Nissim, and Gatt}]{bartl2020unmasking}
Marion Bartl, Malvina Nissim, and Albert Gatt. 2020.
\newblock \href {https://aclanthology.org/2020.gebnlp-1.1} {Unmasking
  contextual stereotypes: Measuring and mitigating {BERT}{'}s gender bias}.
\newblock In \emph{Proceedings of the Second Workshop on Gender Bias in Natural
  Language Processing}, pages 1--16, Barcelona, Spain (Online). Association for
  Computational Linguistics.

\bibitem[{Bentivogli et~al.(2009)Bentivogli, Clark, Dagan, and
  Giampiccolo}]{bentivogli2009fifth}
Luisa Bentivogli, Peter Clark, Ido Dagan, and Danilo Giampiccolo. 2009.
\newblock The fifth pascal recognizing textual entailment challenge.
\newblock In \emph{TAC}.

\bibitem[{Blodgett et~al.(2021)Blodgett, Lopez, Olteanu, Sim, and
  Wallach}]{blodgett-etal-2021-stereotyping}
Su~Lin Blodgett, Gilsinia Lopez, Alexandra Olteanu, Robert Sim, and Hanna
  Wallach. 2021.
\newblock \href {https://doi.org/10.18653/v1/2021.acl-long.81} {Stereotyping
  {N}orwegian salmon: An inventory of pitfalls in fairness benchmark datasets}.
\newblock In \emph{Proceedings of the 59th Annual Meeting of the Association
  for Computational Linguistics and the 11th International Joint Conference on
  Natural Language Processing (Volume 1: Long Papers)}, pages 1004--1015,
  Online. Association for Computational Linguistics.

\bibitem[{Bolukbasi et~al.(2016)Bolukbasi, Chang, Zou, Saligrama, and
  Kalai}]{bolukbasi2016man}
Tolga Bolukbasi, Kai{-}Wei Chang, James~Y. Zou, Venkatesh Saligrama, and
  Adam~Tauman Kalai. 2016.
\newblock \href
  {https://proceedings.neurips.cc/paper/2016/hash/a486cd07e4ac3d270571622f4f316ec5-Abstract.html}
  {Man is to computer programmer as woman is to homemaker? debiasing word
  embeddings}.
\newblock In \emph{Advances in Neural Information Processing Systems 29: Annual
  Conference on Neural Information Processing Systems 2016, December 5-10,
  2016, Barcelona, Spain}, pages 4349--4357.

\bibitem[{Bowman et~al.(2015)Bowman, Angeli, Potts, and
  Manning}]{bowman-etal-2015-large}
Samuel~R. Bowman, Gabor Angeli, Christopher Potts, and Christopher~D. Manning.
  2015.
\newblock \href {https://doi.org/10.18653/v1/D15-1075} {A large annotated
  corpus for learning natural language inference}.
\newblock In \emph{Proceedings of the 2015 Conference on Empirical Methods in
  Natural Language Processing}, pages 632--642, Lisbon, Portugal. Association
  for Computational Linguistics.

\bibitem[{Caliskan et~al.(2017)Caliskan, Bryson, and
  Narayanan}]{caliskan2017semantics}
Aylin Caliskan, Joanna~J Bryson, and Arvind Narayanan. 2017.
\newblock Semantics derived automatically from language corpora contain
  human-like biases.
\newblock \emph{Science}, 356(6334):183--186.

\bibitem[{Cer et~al.(2017)Cer, Diab, Agirre, Lopez-Gazpio, and
  Specia}]{cer2017semeval}
Daniel Cer, Mona Diab, Eneko Agirre, I{\~n}igo Lopez-Gazpio, and Lucia Specia.
  2017.
\newblock \href {https://doi.org/10.18653/v1/S17-2001} {{S}em{E}val-2017 task
  1: Semantic textual similarity multilingual and crosslingual focused
  evaluation}.
\newblock In \emph{Proceedings of the 11th International Workshop on Semantic
  Evaluation ({S}em{E}val-2017)}, pages 1--14, Vancouver, Canada. Association
  for Computational Linguistics.

\bibitem[{Chen et~al.(2020)Chen, Kornblith, Norouzi, and
  Hinton}]{chen2020simple}
Ting Chen, Simon Kornblith, Mohammad Norouzi, and Geoffrey~E. Hinton. 2020.
\newblock \href {http://proceedings.mlr.press/v119/chen20j.html} {A simple
  framework for contrastive learning of visual representations}.
\newblock In \emph{Proceedings of the 37th International Conference on Machine
  Learning, {ICML} 2020, 13-18 July 2020, Virtual Event}, volume 119 of
  \emph{Proceedings of Machine Learning Research}, pages 1597--1607. {PMLR}.

\bibitem[{Cheng et~al.(2021)Cheng, Hao, Yuan, Si, and Carin}]{ChengHYSC21}
Pengyu Cheng, Weituo Hao, Siyang Yuan, Shijing Si, and Lawrence Carin. 2021.
\newblock \href {https://openreview.net/forum?id=N6JECD-PI5w} {Fairfil:
  Contrastive neural debiasing method for pretrained text encoders}.
\newblock In \emph{9th International Conference on Learning Representations,
  {ICLR} 2021, Virtual Event, Austria, May 3-7, 2021}. OpenReview.net.

\bibitem[{Chi et~al.(2022)Chi, Shand, Yu, Chang, Zhao, and
  Tian}]{chi-2022-conditional}
Jianfeng Chi, William Shand, Yaodong Yu, Kai-Wei Chang, Han Zhao, and Yuan
  Tian. 2022.
\newblock \href {https://arxiv.org/abs/2205.11485} {Conditional supervised
  contrastive learning for fair text classification}.

\bibitem[{Conneau and Kiela(2018)}]{conneau2018senteval}
Alexis Conneau and Douwe Kiela. 2018.
\newblock \href {https://aclanthology.org/L18-1269} {{S}ent{E}val: An
  evaluation toolkit for universal sentence representations}.
\newblock In \emph{Proceedings of the Eleventh International Conference on
  Language Resources and Evaluation ({LREC} 2018)}, Miyazaki, Japan. European
  Language Resources Association (ELRA).

\bibitem[{Conneau et~al.(2017)Conneau, Kiela, Schwenk, Barrault, and
  Bordes}]{conneau-etal-2017-supervised}
Alexis Conneau, Douwe Kiela, Holger Schwenk, Lo{\"\i}c Barrault, and Antoine
  Bordes. 2017.
\newblock \href {https://doi.org/10.18653/v1/D17-1070} {Supervised learning of
  universal sentence representations from natural language inference data}.
\newblock In \emph{Proceedings of the 2017 Conference on Empirical Methods in
  Natural Language Processing}, pages 670--680, Copenhagen, Denmark.
  Association for Computational Linguistics.

\bibitem[{Dagan and Glickman(2005)}]{dagan2005pascal}
Ido Dagan and Oren Glickman. 2005.
\newblock The pascal recognising textual entailment challenge.
\newblock In \emph{In Proceedings of the PASCAL Challenges Workshop on
  Recognising Textual Entailment}.

\bibitem[{De-Arteaga et~al.(2019{\natexlab{a}})De-Arteaga, Romanov, Wallach,
  Chayes, Borgs, Chouldechova, Geyik, Kenthapadi, and Kalai}]{de-arteaga-et-al}
Maria De-Arteaga, Alexey Romanov, Hanna Wallach, Jennifer Chayes, Christian
  Borgs, Alexandra Chouldechova, Sahin Geyik, Krishnaram Kenthapadi, and
  Adam~Tauman Kalai. 2019{\natexlab{a}}.
\newblock \href {https://doi.org/10.1145/3287560.3287572} {Bias in bios: A case
  study of semantic representation bias in a high-stakes setting}.
\newblock In \emph{Proceedings of the Conference on Fairness, Accountability,
  and Transparency}, FAT* '19, page 120–128, New York, NY, USA. Association
  for Computing Machinery.

\bibitem[{De-Arteaga et~al.(2019{\natexlab{b}})De-Arteaga, Romanov, Wallach,
  Chayes, Borgs, Chouldechova, Geyik, Kenthapadi, and
  Kalai}]{de-Arteaga2019BiasinBios}
Maria De-Arteaga, Alexey Romanov, Hanna Wallach, Jennifer Chayes, Christian
  Borgs, Alexandra Chouldechova, Sahin Geyik, Krishnaram Kenthapadi, and
  Adam~Tauman Kalai. 2019{\natexlab{b}}.
\newblock \href {https://doi.org/10.1145/3287560.3287572} {Bias in bios: A case
  study of semantic representation bias in a high-stakes setting}.
\newblock In \emph{Proceedings of the Conference on Fairness, Accountability,
  and Transparency}, FAT* '19, page 120–128, New York, NY, USA. Association
  for Computing Machinery.

\bibitem[{Dev et~al.(2020)Dev, Li, Phillips, and
  Srikumar}]{Dev_Li_Phillips_Srikumar_2020}
Sunipa Dev, Tao Li, Jeff~M. Phillips, and Vivek Srikumar. 2020.
\newblock \href {https://ojs.aaai.org/index.php/AAAI/article/view/6267} {On
  measuring and mitigating biased inferences of word embeddings}.
\newblock In \emph{The Thirty-Fourth {AAAI} Conference on Artificial
  Intelligence, {AAAI} 2020, The Thirty-Second Innovative Applications of
  Artificial Intelligence Conference, {IAAI} 2020, The Tenth {AAAI} Symposium
  on Educational Advances in Artificial Intelligence, {EAAI} 2020, New York,
  NY, USA, February 7-12, 2020}, pages 7659--7666. {AAAI} Press.

\bibitem[{Dev et~al.(2021)Dev, Li, Phillips, and Srikumar}]{dev2021oscar}
Sunipa Dev, Tao Li, Jeff~M Phillips, and Vivek Srikumar. 2021.
\newblock \href {https://doi.org/10.18653/v1/2021.emnlp-main.411} {{OSC}a{R}:
  Orthogonal subspace correction and rectification of biases in word
  embeddings}.
\newblock In \emph{Proceedings of the 2021 Conference on Empirical Methods in
  Natural Language Processing}, pages 5034--5050, Online and Punta Cana,
  Dominican Republic. Association for Computational Linguistics.

\bibitem[{Dev and Phillips(2019)}]{pmlr-v89-dev19a}
Sunipa Dev and Jeff~M. Phillips. 2019.
\newblock \href {http://proceedings.mlr.press/v89/dev19a.html} {Attenuating
  bias in word vectors}.
\newblock In \emph{The 22nd International Conference on Artificial Intelligence
  and Statistics, {AISTATS} 2019, 16-18 April 2019, Naha, Okinawa, Japan},
  volume~89 of \emph{Proceedings of Machine Learning Research}, pages 879--887.
  {PMLR}.

\bibitem[{Devlin et~al.(2019)Devlin, Chang, Lee, and
  Toutanova}]{devlin2019bert}
Jacob Devlin, Ming-Wei Chang, Kenton Lee, and Kristina Toutanova. 2019.
\newblock \href {https://doi.org/10.18653/v1/N19-1423} {{BERT}: Pre-training of
  deep bidirectional transformers for language understanding}.
\newblock In \emph{Proceedings of the 2019 Conference of the North {A}merican
  Chapter of the Association for Computational Linguistics: Human Language
  Technologies, Volume 1 (Long and Short Papers)}, pages 4171--4186,
  Minneapolis, Minnesota. Association for Computational Linguistics.

\bibitem[{Dolan and Brockett(2005)}]{dolan2005automatically}
William~B. Dolan and Chris Brockett. 2005.
\newblock \href {https://aclanthology.org/I05-5002} {Automatically constructing
  a corpus of sentential paraphrases}.
\newblock In \emph{Proceedings of the Third International Workshop on
  Paraphrasing ({IWP}2005)}.

\bibitem[{Ethayarajh et~al.(2019)Ethayarajh, Duvenaud, and
  Hirst}]{ethayarajh-etal-2019-understanding}
Kawin Ethayarajh, David Duvenaud, and Graeme Hirst. 2019.
\newblock \href {https://doi.org/10.18653/v1/P19-1166} {Understanding
  undesirable word embedding associations}.
\newblock In \emph{Proceedings of the 57th Annual Meeting of the Association
  for Computational Linguistics}, pages 1696--1705, Florence, Italy.
  Association for Computational Linguistics.

\bibitem[{Gao et~al.(2021)Gao, Yao, and Chen}]{gao2021simcse}
Tianyu Gao, Xingcheng Yao, and Danqi Chen. 2021.
\newblock \href {https://doi.org/10.18653/v1/2021.emnlp-main.552} {{S}im{CSE}:
  Simple contrastive learning of sentence embeddings}.
\newblock In \emph{Proceedings of the 2021 Conference on Empirical Methods in
  Natural Language Processing}, pages 6894--6910, Online and Punta Cana,
  Dominican Republic. Association for Computational Linguistics.

\bibitem[{Giampiccolo et~al.(2008)Giampiccolo, Dang, Magnini, Dagan, Cabrio,
  and Dolan}]{giampiccolo2008fourth}
Danilo Giampiccolo, Hoa~Trang Dang, Bernardo Magnini, Ido Dagan, Elena Cabrio,
  and Bill Dolan. 2008.
\newblock The fourth pascal recognizing textual entailment challenge.
\newblock In \emph{TAC}. Citeseer.

\bibitem[{Giampiccolo et~al.(2007)Giampiccolo, Magnini, Dagan, and
  Dolan}]{giampiccolo2007third}
Danilo Giampiccolo, Bernardo Magnini, Ido Dagan, and Bill Dolan. 2007.
\newblock \href {https://aclanthology.org/W07-1401} {The third {PASCAL}
  recognizing textual entailment challenge}.
\newblock In \emph{Proceedings of the {ACL}-{PASCAL} Workshop on Textual
  Entailment and Paraphrasing}, pages 1--9, Prague. Association for
  Computational Linguistics.

\bibitem[{Goldfarb-Tarrant et~al.(2021)Goldfarb-Tarrant, Marchant,
  Mu{\~n}oz~S{\'a}nchez, Pandya, and
  Lopez}]{goldfarb-tarrant-etal-2021-intrinsic}
Seraphina Goldfarb-Tarrant, Rebecca Marchant, Ricardo Mu{\~n}oz~S{\'a}nchez,
  Mugdha Pandya, and Adam Lopez. 2021.
\newblock \href {https://doi.org/10.18653/v1/2021.acl-long.150} {Intrinsic bias
  metrics do not correlate with application bias}.
\newblock In \emph{Proceedings of the 59th Annual Meeting of the Association
  for Computational Linguistics and the 11th International Joint Conference on
  Natural Language Processing (Volume 1: Long Papers)}, pages 1926--1940,
  Online. Association for Computational Linguistics.

\bibitem[{Guo and Caliskan(2021)}]{10.1145/3461702.3462536}
Wei Guo and Aylin Caliskan. 2021.
\newblock \href {https://doi.org/10.1145/3461702.3462536} {\emph{Detecting
  Emergent Intersectional Biases: Contextualized Word Embeddings Contain a
  Distribution of Human-like Biases}}, page 122–133. Association for
  Computing Machinery, New York, NY, USA.

\bibitem[{Guo et~al.(2022)Guo, Yang, and Abbasi}]{guo-etal-2022-auto}
Yue Guo, Yi~Yang, and Ahmed Abbasi. 2022.
\newblock \href {https://doi.org/10.18653/v1/2022.acl-long.72} {Auto-debias:
  Debiasing masked language models with automated biased prompts}.
\newblock In \emph{Proceedings of the 60th Annual Meeting of the Association
  for Computational Linguistics (Volume 1: Long Papers)}, pages 1012--1023,
  Dublin, Ireland. Association for Computational Linguistics.

\bibitem[{Haim et~al.(2006)Haim, Dagan, Dolan, Ferro, Giampiccolo, Magnini, and
  Szpektor}]{haim2006second}
R~Bar Haim, Ido Dagan, Bill Dolan, Lisa Ferro, Danilo Giampiccolo, Bernardo
  Magnini, and Idan Szpektor. 2006.
\newblock The second pascal recognising textual entailment challenge.
\newblock In \emph{Proceedings of the Second PASCAL Challenges Workshop on
  Recognising Textual Entailment}, volume~7.

\bibitem[{Han et~al.(2021{\natexlab{a}})Han, Baldwin, and
  Cohn}]{han2021balancing}
Xudong Han, Timothy Baldwin, and Trevor Cohn. 2021{\natexlab{a}}.
\newblock \href {https://arxiv.org/abs/2109.08253} {Balancing out bias:
  Achieving fairness through training reweighting}.
\newblock \emph{ArXiv preprint}, abs/2109.08253.

\bibitem[{Han et~al.(2021{\natexlab{b}})Han, Baldwin, and
  Cohn}]{han2021diverse}
Xudong Han, Timothy Baldwin, and Trevor Cohn. 2021{\natexlab{b}}.
\newblock \href {https://doi.org/10.18653/v1/2021.eacl-main.239} {Diverse
  adversaries for mitigating bias in training}.
\newblock In \emph{Proceedings of the 16th Conference of the European Chapter
  of the Association for Computational Linguistics: Main Volume}, pages
  2760--2765, Online. Association for Computational Linguistics.

\bibitem[{Hovy et~al.(2006)Hovy, Marcus, Palmer, Ramshaw, and
  Weischedel}]{hovy-etal-2006-ontonotes}
Eduard Hovy, Mitchell Marcus, Martha Palmer, Lance Ramshaw, and Ralph
  Weischedel. 2006.
\newblock \href {https://aclanthology.org/N06-2015} {{O}nto{N}otes: The 90{\%}
  solution}.
\newblock In \emph{Proceedings of the Human Language Technology Conference of
  the {NAACL}, Companion Volume: Short Papers}, pages 57--60, New York City,
  USA. Association for Computational Linguistics.

\bibitem[{Jin et~al.(2021)Jin, Barbieri, Kennedy, Mostafazadeh~Davani, Neves,
  and Ren}]{jin-etal-2021-transferability}
Xisen Jin, Francesco Barbieri, Brendan Kennedy, Aida Mostafazadeh~Davani,
  Leonardo Neves, and Xiang Ren. 2021.
\newblock \href {https://doi.org/10.18653/v1/2021.naacl-main.296} {On
  transferability of bias mitigation effects in language model fine-tuning}.
\newblock In \emph{Proceedings of the 2021 Conference of the North American
  Chapter of the Association for Computational Linguistics: Human Language
  Technologies}, pages 3770--3783, Online. Association for Computational
  Linguistics.

\bibitem[{Kaneko and Bollegala(2021)}]{kaneko-bollegala-2021-context}
Masahiro Kaneko and Danushka Bollegala. 2021.
\newblock \href {https://doi.org/10.18653/v1/2021.eacl-main.107} {Debiasing
  pre-trained contextualised embeddings}.
\newblock In \emph{Proceedings of the 16th Conference of the European Chapter
  of the Association for Computational Linguistics: Main Volume}, pages
  1256--1266, Online. Association for Computational Linguistics.

\bibitem[{Kaneko et~al.(2022)Kaneko, Bollegala, and
  Okazaki}]{kaneko2022debiasing}
Masahiro Kaneko, Danushka Bollegala, and Naoaki Okazaki. 2022.
\newblock \href {https://aclanthology.org/2022.coling-1.111} {Debiasing isn{'}t
  enough! {--} on the effectiveness of debiasing {MLM}s and their social biases
  in downstream tasks}.
\newblock In \emph{Proceedings of the 29th International Conference on
  Computational Linguistics}, pages 1299--1310, Gyeongju, Republic of Korea.
  International Committee on Computational Linguistics.

\bibitem[{Kurita et~al.(2019)Kurita, Vyas, Pareek, Black, and
  Tsvetkov}]{kurita-etal-2019-measuring}
Keita Kurita, Nidhi Vyas, Ayush Pareek, Alan~W Black, and Yulia Tsvetkov. 2019.
\newblock \href {https://doi.org/10.18653/v1/W19-3823} {Measuring bias in
  contextualized word representations}.
\newblock In \emph{Proceedings of the First Workshop on Gender Bias in Natural
  Language Processing}, pages 166--172, Florence, Italy. Association for
  Computational Linguistics.

\bibitem[{Lauscher et~al.(2021)Lauscher, Lueken, and
  Glava{\v{s}}}]{lauscher-etal-2021-sustainable-modular}
Anne Lauscher, Tobias Lueken, and Goran Glava{\v{s}}. 2021.
\newblock \href {https://doi.org/10.18653/v1/2021.findings-emnlp.411}
  {Sustainable modular debiasing of language models}.
\newblock In \emph{Findings of the Association for Computational Linguistics:
  EMNLP 2021}, pages 4782--4797, Punta Cana, Dominican Republic. Association
  for Computational Linguistics.

\bibitem[{Liang et~al.(2020)Liang, Li, Zheng, Lim, Salakhutdinov, and
  Morency}]{liang-etal-2020-towards}
Paul~Pu Liang, Irene~Mengze Li, Emily Zheng, Yao~Chong Lim, Ruslan
  Salakhutdinov, and Louis-Philippe Morency. 2020.
\newblock \href {https://doi.org/10.18653/v1/2020.acl-main.488} {Towards
  debiasing sentence representations}.
\newblock In \emph{Proceedings of the 58th Annual Meeting of the Association
  for Computational Linguistics}, pages 5502--5515, Online. Association for
  Computational Linguistics.

\bibitem[{Liu et~al.(2019)Liu, Ott, Goyal, Du, Joshi, Chen, Levy, Lewis,
  Zettlemoyer, and Stoyanov}]{liu2019roberta}
Yinhan Liu, Myle Ott, Naman Goyal, Jingfei Du, Mandar Joshi, Danqi Chen, Omer
  Levy, Mike Lewis, Luke Zettlemoyer, and Veselin Stoyanov. 2019.
\newblock \href {https://arxiv.org/abs/1907.11692} {{RoBERTa}: {A} robustly
  optimized {BERT} pretraining approach}.
\newblock \emph{ArXiv preprint}, abs/1907.11692.

\bibitem[{May et~al.(2019)May, Wang, Bordia, Bowman, and
  Rudinger}]{may-etal-2019-measuring}
Chandler May, Alex Wang, Shikha Bordia, Samuel~R. Bowman, and Rachel Rudinger.
  2019.
\newblock \href {https://doi.org/10.18653/v1/N19-1063} {On measuring social
  biases in sentence encoders}.
\newblock In \emph{Proceedings of the 2019 Conference of the North {A}merican
  Chapter of the Association for Computational Linguistics: Human Language
  Technologies, Volume 1 (Long and Short Papers)}, pages 622--628, Minneapolis,
  Minnesota. Association for Computational Linguistics.

\bibitem[{Meade et~al.(2022)Meade, Poole-Dayan, and Reddy}]{meadeempirical2022}
Nicholas Meade, Elinor Poole-Dayan, and Siva Reddy. 2022.
\newblock \href {https://doi.org/10.18653/v1/2022.acl-long.132} {An empirical
  survey of the effectiveness of debiasing techniques for pre-trained language
  models}.
\newblock In \emph{Proceedings of the 60th Annual Meeting of the Association
  for Computational Linguistics (Volume 1: Long Papers)}, pages 1878--1898,
  Dublin, Ireland. Association for Computational Linguistics.

\bibitem[{Nadeem et~al.(2021)Nadeem, Bethke, and
  Reddy}]{nadeem-etal-2021-stereoset}
Moin Nadeem, Anna Bethke, and Siva Reddy. 2021.
\newblock \href {https://doi.org/10.18653/v1/2021.acl-long.416} {{S}tereo{S}et:
  Measuring stereotypical bias in pretrained language models}.
\newblock In \emph{Proceedings of the 59th Annual Meeting of the Association
  for Computational Linguistics and the 11th International Joint Conference on
  Natural Language Processing (Volume 1: Long Papers)}, pages 5356--5371,
  Online. Association for Computational Linguistics.

\bibitem[{Nangia et~al.(2020)Nangia, Vania, Bhalerao, and
  Bowman}]{nangia-etal-2020-crows}
Nikita Nangia, Clara Vania, Rasika Bhalerao, and Samuel~R. Bowman. 2020.
\newblock \href {https://doi.org/10.18653/v1/2020.emnlp-main.154}
  {{C}row{S}-pairs: A challenge dataset for measuring social biases in masked
  language models}.
\newblock In \emph{Proceedings of the 2020 Conference on Empirical Methods in
  Natural Language Processing (EMNLP)}, pages 1953--1967, Online. Association
  for Computational Linguistics.

\bibitem[{Orgad et~al.(2022)Orgad, Goldfarb-Tarrant, and
  Belinkov}]{orgad2022how}
Hadas Orgad, Seraphina Goldfarb-Tarrant, and Yonatan Belinkov. 2022.
\newblock \href {https://doi.org/10.18653/v1/2022.naacl-main.188} {How gender
  debiasing affects internal model representations, and why it matters}.
\newblock In \emph{Proceedings of the 2022 Conference of the North American
  Chapter of the Association for Computational Linguistics: Human Language
  Technologies}, pages 2602--2628, Seattle, United States. Association for
  Computational Linguistics.

\bibitem[{Paszke et~al.(2019)Paszke, Gross, Massa, Lerer, Bradbury, Chanan,
  Killeen, Lin, Gimelshein, Antiga, Desmaison, K{\"{o}}pf, Yang, DeVito,
  Raison, Tejani, Chilamkurthy, Steiner, Fang, Bai, and
  Chintala}]{NEURIPS2019_9015}
Adam Paszke, Sam Gross, Francisco Massa, Adam Lerer, James Bradbury, Gregory
  Chanan, Trevor Killeen, Zeming Lin, Natalia Gimelshein, Luca Antiga, Alban
  Desmaison, Andreas K{\"{o}}pf, Edward Yang, Zachary DeVito, Martin Raison,
  Alykhan Tejani, Sasank Chilamkurthy, Benoit Steiner, Lu~Fang, Junjie Bai, and
  Soumith Chintala. 2019.
\newblock \href
  {https://proceedings.neurips.cc/paper/2019/hash/bdbca288fee7f92f2bfa9f7012727740-Abstract.html}
  {Pytorch: An imperative style, high-performance deep learning library}.
\newblock In \emph{Advances in Neural Information Processing Systems 32: Annual
  Conference on Neural Information Processing Systems 2019, NeurIPS 2019,
  December 8-14, 2019, Vancouver, BC, Canada}, pages 8024--8035.

\bibitem[{Peters et~al.(2018)Peters, Neumann, Iyyer, Gardner, Clark, Lee, and
  Zettlemoyer}]{peters-etal-2018-deep}
Matthew~E. Peters, Mark Neumann, Mohit Iyyer, Matt Gardner, Christopher Clark,
  Kenton Lee, and Luke Zettlemoyer. 2018.
\newblock \href {https://doi.org/10.18653/v1/N18-1202} {Deep contextualized
  word representations}.
\newblock In \emph{Proceedings of the 2018 Conference of the North {A}merican
  Chapter of the Association for Computational Linguistics: Human Language
  Technologies, Volume 1 (Long Papers)}, pages 2227--2237, New Orleans,
  Louisiana. Association for Computational Linguistics.

\bibitem[{Qian et~al.(2022)Qian, Ross, Fernandes, Smith, Kiela, and
  Williams}]{qian2022perturbation}
Rebecca Qian, Candace Ross, Jude Fernandes, Eric Smith, Douwe Kiela, and Adina
  Williams. 2022.
\newblock \href {https://arxiv.org/abs/2205.12586} {Perturbation augmentation
  for fairer nlp}.

\bibitem[{Rajpurkar et~al.(2016)Rajpurkar, Zhang, Lopyrev, and
  Liang}]{rajpurkar-etal-2016-squad}
Pranav Rajpurkar, Jian Zhang, Konstantin Lopyrev, and Percy Liang. 2016.
\newblock \href {https://doi.org/10.18653/v1/D16-1264} {{SQ}u{AD}: 100,000+
  questions for machine comprehension of text}.
\newblock In \emph{Proceedings of the 2016 Conference on Empirical Methods in
  Natural Language Processing}, pages 2383--2392, Austin, Texas. Association
  for Computational Linguistics.

\bibitem[{Ravfogel et~al.(2020)Ravfogel, Elazar, Gonen, Twiton, and
  Goldberg}]{ravfogel-etal-2020-null}
Shauli Ravfogel, Yanai Elazar, Hila Gonen, Michael Twiton, and Yoav Goldberg.
  2020.
\newblock \href {https://doi.org/10.18653/v1/2020.acl-main.647} {Null it out:
  Guarding protected attributes by iterative nullspace projection}.
\newblock In \emph{Proceedings of the 58th Annual Meeting of the Association
  for Computational Linguistics}, pages 7237--7256, Online. Association for
  Computational Linguistics.

\bibitem[{Ravfogel et~al.(2022)Ravfogel, Twiton, Goldberg, and
  Cotterell}]{ravfogel2022linear}
Shauli Ravfogel, Michael Twiton, Yoav Goldberg, and Ryan Cotterell. 2022.
\newblock Linear adversarial concept erasure.
\newblock In \emph{International Conference on Machine Learning}. PMLR.

\bibitem[{Reimers and Gurevych(2019)}]{reimers2019sentence}
Nils Reimers and Iryna Gurevych. 2019.
\newblock \href {https://doi.org/10.18653/v1/D19-1410} {Sentence-{BERT}:
  Sentence embeddings using {S}iamese {BERT}-networks}.
\newblock In \emph{Proceedings of the 2019 Conference on Empirical Methods in
  Natural Language Processing and the 9th International Joint Conference on
  Natural Language Processing (EMNLP-IJCNLP)}, pages 3982--3992, Hong Kong,
  China. Association for Computational Linguistics.

\bibitem[{Rudinger et~al.(2018)Rudinger, Naradowsky, Leonard, and
  Van~Durme}]{rudinger-EtAl:2018:N18}
Rachel Rudinger, Jason Naradowsky, Brian Leonard, and Benjamin Van~Durme. 2018.
\newblock \href {https://doi.org/10.18653/v1/N18-2002} {Gender bias in
  coreference resolution}.
\newblock In \emph{Proceedings of the 2018 Conference of the North {A}merican
  Chapter of the Association for Computational Linguistics: Human Language
  Technologies, Volume 2 (Short Papers)}, pages 8--14, New Orleans, Louisiana.
  Association for Computational Linguistics.

\bibitem[{Shen et~al.(2021)Shen, Han, Cohn, Baldwin, and
  Frermann}]{shen-2021-contrastive}
Aili Shen, Xudong Han, Trevor Cohn, Timothy Baldwin, and Lea Frermann. 2021.
\newblock \href {https://arxiv.org/abs/2109.10645} {Contrastive learning for
  fair representations}.

\bibitem[{Silva et~al.(2021)Silva, Tambwekar, and
  Gombolay}]{silva-etal-2021-towards}
Andrew Silva, Pradyumna Tambwekar, and Matthew Gombolay. 2021.
\newblock \href {https://doi.org/10.18653/v1/2021.naacl-main.189} {Towards a
  comprehensive understanding and accurate evaluation of societal biases in
  pre-trained transformers}.
\newblock In \emph{Proceedings of the 2021 Conference of the North American
  Chapter of the Association for Computational Linguistics: Human Language
  Technologies}, pages 2383--2389, Online. Association for Computational
  Linguistics.

\bibitem[{Socher et~al.(2013)Socher, Perelygin, Wu, Chuang, Manning, Ng, and
  Potts}]{socher2013recursive_sst-2}
Richard Socher, Alex Perelygin, Jean Wu, Jason Chuang, Christopher~D. Manning,
  Andrew Ng, and Christopher Potts. 2013.
\newblock \href {https://aclanthology.org/D13-1170} {Recursive deep models for
  semantic compositionality over a sentiment treebank}.
\newblock In \emph{Proceedings of the 2013 Conference on Empirical Methods in
  Natural Language Processing}, pages 1631--1642, Seattle, Washington, USA.
  Association for Computational Linguistics.

\bibitem[{Subramanian et~al.(2021)Subramanian, Han, Baldwin, Cohn, and
  Frermann}]{subramanian-etal-2021-evaluating}
Shivashankar Subramanian, Xudong Han, Timothy Baldwin, Trevor Cohn, and Lea
  Frermann. 2021.
\newblock \href {https://doi.org/10.18653/v1/2021.emnlp-main.193} {Evaluating
  debiasing techniques for intersectional biases}.
\newblock In \emph{Proceedings of the 2021 Conference on Empirical Methods in
  Natural Language Processing}, pages 2492--2498, Online and Punta Cana,
  Dominican Republic. Association for Computational Linguistics.

\bibitem[{van~der Maaten and Hinton(2008)}]{JMLR:v9:vandermaaten08a}
Laurens van~der Maaten and Geoffrey Hinton. 2008.
\newblock \href {http://jmlr.org/papers/v9/vandermaaten08a.html} {Visualizing
  data using t-sne}.
\newblock \emph{Journal of Machine Learning Research}, 9(86):2579--2605.

\bibitem[{Wang et~al.(2019)Wang, Singh, Michael, Hill, Levy, and
  Bowman}]{wang-etal-2018-glue}
Alex Wang, Amanpreet Singh, Julian Michael, Felix Hill, Omer Levy, and
  Samuel~R. Bowman. 2019.
\newblock \href {https://openreview.net/forum?id=rJ4km2R5t7} {{GLUE:} {A}
  multi-task benchmark and analysis platform for natural language
  understanding}.
\newblock In \emph{7th International Conference on Learning Representations,
  {ICLR} 2019, New Orleans, LA, USA, May 6-9, 2019}. OpenReview.net.

\bibitem[{Wang and Isola(2020)}]{wang2020hypersphere}
Tongzhou Wang and Phillip Isola. 2020.
\newblock \href {http://proceedings.mlr.press/v119/wang20k.html} {Understanding
  contrastive representation learning through alignment and uniformity on the
  hypersphere}.
\newblock In \emph{Proceedings of the 37th International Conference on Machine
  Learning, {ICML} 2020, 13-18 July 2020, Virtual Event}, volume 119 of
  \emph{Proceedings of Machine Learning Research}, pages 9929--9939. {PMLR}.

\bibitem[{Wang et~al.(2017)Wang, Hamza, and Florian}]{wang2017bilateral}
Zhiguo Wang, Wael Hamza, and Radu Florian. 2017.
\newblock \href {https://doi.org/10.24963/ijcai.2017/579} {Bilateral
  multi-perspective matching for natural language sentences}.
\newblock In \emph{Proceedings of the Twenty-Sixth International Joint
  Conference on Artificial Intelligence, {IJCAI} 2017, Melbourne, Australia,
  August 19-25, 2017}, pages 4144--4150. ijcai.org.

\bibitem[{Warstadt et~al.(2019)Warstadt, Singh, and
  Bowman}]{warstadt2019neural}
Alex Warstadt, Amanpreet Singh, and Samuel~R. Bowman. 2019.
\newblock \href {https://doi.org/10.1162/tacl_a_00290} {Neural network
  acceptability judgments}.
\newblock \emph{Transactions of the Association for Computational Linguistics},
  7:625--641.

\bibitem[{Webster et~al.(2020)Webster, Wang, Tenney, Beutel, Pitler, Pavlick,
  Chen, Chi, and Petrov}]{webster2020measuring}
Kellie Webster, Xuezhi Wang, Ian Tenney, Alex Beutel, Emily Pitler, Ellie
  Pavlick, Jilin Chen, Ed~Chi, and Slav Petrov. 2020.
\newblock \href {https://arxiv.org/abs/2010.06032} {Measuring and reducing
  gendered correlations in pre-trained models}.

\bibitem[{Wieting and Gimpel(2018)}]{wieting-gimpel-2018-paranmt}
John Wieting and Kevin Gimpel. 2018.
\newblock \href {https://doi.org/10.18653/v1/P18-1042} {{P}ara{NMT}-50{M}:
  Pushing the limits of paraphrastic sentence embeddings with millions of
  machine translations}.
\newblock In \emph{Proceedings of the 56th Annual Meeting of the Association
  for Computational Linguistics (Volume 1: Long Papers)}, pages 451--462,
  Melbourne, Australia. Association for Computational Linguistics.

\bibitem[{Williams et~al.(2018)Williams, Nangia, and Bowman}]{N18-1101}
Adina Williams, Nikita Nangia, and Samuel Bowman. 2018.
\newblock \href {https://doi.org/10.18653/v1/N18-1101} {A broad-coverage
  challenge corpus for sentence understanding through inference}.
\newblock In \emph{Proceedings of the 2018 Conference of the North {A}merican
  Chapter of the Association for Computational Linguistics: Human Language
  Technologies, Volume 1 (Long Papers)}, pages 1112--1122, New Orleans,
  Louisiana. Association for Computational Linguistics.

\bibitem[{Wolf et~al.(2020)Wolf, Debut, Sanh, Chaumond, Delangue, Moi, Cistac,
  Rault, Louf, Funtowicz, Davison, Shleifer, von Platen, Ma, Jernite, Plu, Xu,
  Le~Scao, Gugger, Drame, Lhoest, and Rush}]{wolf-etal-2020-transformers}
Thomas Wolf, Lysandre Debut, Victor Sanh, Julien Chaumond, Clement Delangue,
  Anthony Moi, Pierric Cistac, Tim Rault, Remi Louf, Morgan Funtowicz, Joe
  Davison, Sam Shleifer, Patrick von Platen, Clara Ma, Yacine Jernite, Julien
  Plu, Canwen Xu, Teven Le~Scao, Sylvain Gugger, Mariama Drame, Quentin Lhoest,
  and Alexander Rush. 2020.
\newblock \href {https://doi.org/10.18653/v1/2020.emnlp-demos.6} {Transformers:
  State-of-the-art natural language processing}.
\newblock In \emph{Proceedings of the 2020 Conference on Empirical Methods in
  Natural Language Processing: System Demonstrations}, pages 38--45, Online.
  Association for Computational Linguistics.

\bibitem[{Xu and Choi(2020)}]{xu-choi-2020-revealing}
Liyan Xu and Jinho~D. Choi. 2020.
\newblock \href {https://doi.org/10.18653/v1/2020.emnlp-main.686} {Revealing
  the myth of higher-order inference in coreference resolution}.
\newblock In \emph{Proceedings of the 2020 Conference on Empirical Methods in
  Natural Language Processing (EMNLP)}, pages 8527--8533, Online. Association
  for Computational Linguistics.

\bibitem[{Zhao et~al.(2019)Zhao, Wang, Yatskar, Cotterell, Ordonez, and
  Chang}]{zhao-etal-2019-gender}
Jieyu Zhao, Tianlu Wang, Mark Yatskar, Ryan Cotterell, Vicente Ordonez, and
  Kai-Wei Chang. 2019.
\newblock \href {https://doi.org/10.18653/v1/N19-1064} {Gender bias in
  contextualized word embeddings}.
\newblock In \emph{Proceedings of the 2019 Conference of the North {A}merican
  Chapter of the Association for Computational Linguistics: Human Language
  Technologies, Volume 1 (Long and Short Papers)}, pages 629--634, Minneapolis,
  Minnesota. Association for Computational Linguistics.

\bibitem[{Zhao et~al.(2018)Zhao, Wang, Yatskar, Ordonez, and
  Chang}]{zhao-etal-2018-gender}
Jieyu Zhao, Tianlu Wang, Mark Yatskar, Vicente Ordonez, and Kai-Wei Chang.
  2018.
\newblock \href {https://doi.org/10.18653/v1/N18-2003} {Gender bias in
  coreference resolution: Evaluation and debiasing methods}.
\newblock In \emph{Proceedings of the 2018 Conference of the North {A}merican
  Chapter of the Association for Computational Linguistics: Human Language
  Technologies, Volume 2 (Short Papers)}, pages 15--20, New Orleans, Louisiana.
  Association for Computational Linguistics.

\end{thebibliography}
\bibliographystyle{acl_natbib}

\appendix 

\clearpage
\section{Implementation Details of {\ours}}
\label{sec:appendix_method_impl}
 We use an aggregation of entailment pairs from the SNLI and MNLI datasets, and augment pairs with opposite gender directions, drawing from the same list of attribute word pairs used by \citet{bolukbasi2016man}, \citet{liang-etal-2020-towards}, and \citet{ChengHYSC21}: \texttt{(man, woman), (boy, girl), (he, she), (father, mother), (son, daughter), (guy, gal), (male, female), (his, her), (himself, herself), (John, Mary)}, alongside plural forms.

We implement \ours{} using the HuggingFace Trainer in PyTorch \cite{NEURIPS2019_9015} and train for 2 epochs. We take the last-saved checkpoint. Training \ours{} takes less than 2 hours across 4 NVIDIA GeForce RTX 3090 GPUs.

\paragraph{Ablation details.}
Dataset sizes from our ablation study are in \autoref{tab:app_data_info}. 
\begin{table}[!ht]
\centering
\resizebox{0.47\textwidth}{!}{
\begin{tabular}{lccc}
\toprule
\tf{Dataset} & \tf{Type} & \tf{Original \#}  & \tf{Final \#}  \\
\midrule
\textsc{MNLI}  & Entailment & 130.9K & 21.5K \\
\textsc{SNLI} & Entailment & 190.1K & 112.7K \\
\textsc{SNLI} & Neutral & 189.2K & 126.6K \\
\textsc{SNLI} & Contradiction & 189.7K & 127.2K \\
\textsc{QQP} & Paraphrase & 149.2K & 23.9K \\
\textsc{Para-NMT} & Paraphrase & 5M & 1.3M \\
\bottomrule
\end{tabular}}
\caption{Information about dataset sizes. }
\label{tab:app_data_info}
\end{table}


Besides batch size, we also tune for learning rate $\in$ \{$1e^{-5}$, $3e^{-5}$, $5e^{-5}$\} and $\alpha \in$ \{0.01, 0.05, 0.1\}. 
 
As \autoref{tab:app_lr_alpha_tuning} indicates, increasing the learning rate improves fairness as the stereotype score approaches 50, but also seems to slightly diminish the model's language modeling ability. Furthermore, a larger $\alpha$ results in a fairer stereotype score, which corroborates our intuition as this parameter adjusts the influence of our alignment loss. Unfortunately, increasing $\alpha$ also monotonically decreases the language modeling score.  

Therefore, we take a learning rate of $5e^{-5}$, a batch size of 32, and an $\alpha = 0.05$ as our default hyper-parameters; these result in the best trade-off between fairness and language modeling ability. We use $\lambda = 0.1$ in all the experiments. 

\begin{table}[ht]
\centering
\begin{tabular}{@{}lccc@{}}
\toprule
  & \multicolumn{1}{c}{LM  $\uparrow$} & \multicolumn{1}{c}{SS $\diamond$}& \multicolumn{1}{c}{ICAT  $\uparrow$} \\ \midrule
LR~$=1e^{-5}$ & 85.13 & 59.71  & 68.60\\
LR~$=3e^{-5}$ & 85.03 & 58.29  & 70.92\\
LR~$=5e^{-5}$  & 84.54 & \bf{56.73} & \bf{73.98}\\  \midrule 
$\alpha=0.01$  & \bf{85.29} & 59.67 & 68.80\\
$\alpha=0.05$ & 84.54 & \bf{56.73} & \bf{73.98}\\
$\alpha=0.1$  & 83.34 & 56.94 & 72.52\\ \bottomrule
\end{tabular}
\caption{StereoSet results on different hyper-parameter settings. Unless otherwise stated, the default configuration is a learning rate (LR) of $5e^{-5}$, a batch size of 32, and an $\alpha$ of 0.05. $\diamond$: the closer to 50, the better.}
\label{tab:app_lr_alpha_tuning}
\end{table} 

\section{Baseline Implementation}
\label{sec:appendix_baseline_impl}

\paragraph{Context-Debias.}
We use the model checkpoint provided by \citet{kaneko-bollegala-2021-context}, and treat it as a regular encoder for downstream evaluation. 

\paragraph{Sent-Debias.} We use the code and data provided by \citet{liang-etal-2020-towards} to compute the gender bias subspace. For downstream evaluation, the debiasing step (subtracting the subspace from the representations) is applied directly after encoding.

\paragraph{\ff{}.}
As code for this work is not available, we re-implement \ff{}, the main contrastive approach, without the additional information-theoretic regularizer. Note that the reported performance difference from including the regularizer or not (0.150 vs. 0.179 on SEAT) is marginal. We checked all the implementation details carefully and report our reproduced and original SEAT effect size results in \autoref{tab:app_seat_fairfil}.

\begin{table}[ht]
\begin{tabular}{@{}lcc@{}}
\toprule 
\multicolumn{1}{l}{\tf{SEAT}} & \multicolumn{1}{l}{\tf{FF (O)}} & \multicolumn{1}{l}{\tf{FF (R)}}   \\
\tf{Category} &  \tf{ES} $\ddagger$ & \tf{ES} $\ddagger$ \\ \midrule
Names, Career/Family 6  & 0.218 & 0.279\small{±0.147}  \\
Terms, Career/Family 6b  & 0.086 & 0.155\small{±0.139}  \\
Terms, Math/Arts 7  & 0.133 & 0.046\small{±0.008}   \\
Names, Math/Arts 7b  & 0.101 & 0.061\small{±0.046}   \\
Terms, Science/Arts 8  & 0.218 & 0.055\small{±0.050}   \\
Names, Science/Arts 8b  & 0.320 & 0.530\small{±0.092}   \\
Avg. Abs. Effect Size & 0.179 & 0.188  \\ \bottomrule
\end{tabular}
\caption{Absolute average effect sizes (ES) on the 6 gender-associated SEAT categories, for original (O) and reproduced (R) results on \ff{} (FF). We report the average and standard deviation for our reproduction. $\ddagger$: the closer to 0, the better.}
\label{tab:app_seat_fairfil}
\end{table}

During evaluation, we fix the \ff{} layer upon initialization so that its parameters no longer update. We debias by feeding encoded representations through the layer. 

\section{Evaluation Details}
\label{sec:appendix_evaluation}

\subsection{Intrinsic Metrics}

\paragraph{StereoSet.}
StereoSet unifies the language modeling (LM) score and the stereotype score (SS) into a single metric, the Idealized Context Association Test (ICAT) score, which is as follows:

\begin{align*}
    \textrm{ICAT} = \textrm{LM} \cdot \frac{\min(\textrm{SS}, 100 - \textrm{SS})}{50}.
\end{align*}

In the ideal scenario, a perfectly fair and highly performative language model would have an LM score of 100, an SS score of 50, and thus an ICAT score of 100. Therefore, the higher the ICAT score, the better.

\paragraph{CrowS-Pairs.}
While CrowS-Pairs originally used pseudo log-likelihood MLM scoring, this form of measurement is found to be error-prone \cite{meadeempirical2022}. Therefore, we follow \citet{meadeempirical2022}'s evaluation approach, and compare the masked token probability of tokens unique to each sentence. The stereotype score (SS) for this task is the percentage of instances for which a language model computes a greater masked token probability to a stereotypical sentence over an anti-stereotypical sentence. An impartial language model without stereotypical biases should score an SS of 50.

\subsection{Extrinsic Metrics}

\paragraph{Bias-in-Bios.} 
$GAP^{TPR}_M$ is denoted as (observe that the closer the value is to 0, the better)

\begin{equation*}
    GAP^{TPR}_M = |TPR_M - TPR_F|.
    \label{eq:tpr}
\end{equation*}

Merely taking the difference in overall accuracies does not account for the highly imbalanced nature of the Bias-in-Bios dataset. In line with \citet{ravfogel-etal-2020-null}, we also calculate the root-mean square of $GAP^{TPR}_{M, y}$ to obtain a more robust metric. Taking $y$ as a profession in $C$, the set of all 28 professions, we can compute

\begin{equation*}
    GAP^{TPR, RMS}_M = \sqrt{\frac{1}{|C|} \sum_{y \in C}(GAP^{TPR}_{M, y})^2}.
\end{equation*}


\begin{table*}[t]
\centering
\small
\begin{tabular}{p{0.13\linewidth}p{0.32\linewidth}p{0.32\linewidth}}
\toprule
 & \tf{Type 1 Sentence} & \tf{Type 2 Sentence} \\ \midrule
Pro-stereotypical & \colorbox{red!20}{The developer} argued with the designer because \colorbox{red!20}{he} did not like the design. & The guard admired \colorbox{red!20}{the secretary} and wanted \colorbox{red!20}{her} job. \\ \midrule
Anti-stereotypical & The developer argued with \colorbox{blue!20}{the designer} because \colorbox{blue!20}{his} design cannot be implemented. & The secretary called \colorbox{blue!20}{the mover} and asked \colorbox{blue!20}{her} to come. \\ \bottomrule
\end{tabular}
    \caption{Example of Type I and Type II sentences from the WinoBias dataset \cite{zhao-etal-2018-gender}. The colored text indicates the pronoun and the correct coreferent. 
    }
    \label{tab:winobias}
\end{table*}

Following the suggestion of \citet{de-arteaga-et-al}, our train-val-test split of the Bias-in-Bios dataset is 65/25/10. We were able to scrape 206,511 biographies.\footnote{\url{https://github.com/microsoft/biosbias}}

In the fine-tuning setting, we train for 5 epochs and evaluate every 1000 steps on the validation set. The model checkpoint is saved if the validation accuracy has improved.
We use the \texttt{AutoModelForSequenceClassification} class from the \texttt{transformers} package \cite{wolf-etal-2020-transformers}, which extracts sentence representations by taking the last-layer hidden state of the \texttt{[CLS]} token and feeding it through a linear layer with \textit{tanh} activation. We use a batch size of 128, a learning rate of $\lambda = 1e^{-5}$, and a maximum sequence length of 128.

\paragraph{Bias-NLI.} 
The three evaluation metrics used in Bias-NLI task are calculated as follows:

\begin{enumerate}
    \item \textbf{Net Neutral (NN)}: The average probability of the neutral label across all instances.
    \item \textbf{Fraction Neutral (FN)}: The fraction of sentence pairs accurately labeled as neutral.
    \item \textbf{Threshold:$\tau$ (T:$\tau$)}: The fraction of instances with the probability of neutral above $\tau$.
\end{enumerate}


In the linear probing setting, we construct an updating linear layer on top of the frozen encoder. Following \citet{Dev_Li_Phillips_Srikumar_2020}, sentence representations are extracted from the \texttt{[CLS]} token of the last hidden state. We use a batch size of 64, a learning rate of  $\lambda = 5 \times 10^{-5}$, and a maximum sequence length of 128. We fine-tune for 3 epochs and evaluate every 500 steps, saving the checkpoint if the validation accuracy improves. We randomly sub-sample 10,000 elements from \citet{Dev_Li_Phillips_Srikumar_2020}'s evaluation dataset during inference.

\paragraph{WinoBias.} Each WinoBias example contains exactly two mentions of professions and one pronoun, which co-refers correctly to one of the profession (\autoref{tab:winobias}). Type 1 sentences are syntactically ambiguous and require world knowledge to be correctly resolved, while Type 2 sentences are easier and can be inferred through only syntactic cues. Examples are presented in in \autoref{tab:winobias}.

Following previous gender bias analyses \cite{orgad2022how}, we borrow the PyTorch re-implementation of the end-to-end c2f-coref model from \citet{xu-choi-2020-revealing}. We use the \textit{cased} version of encoders, a significant performance difference exists between cased and uncased variants. Models are trained for 24 epochs with a dropout rate of $0.3$ and a maximum sequence length of 384. Encoder parameters and task parameters have separate learning rates ($1 \times 10^{-5}$ and $3 \times 10^{-4}$), separate linear decay schedules, and separate weight decay rates ($1 \times 10^{-2}$ and $0$).

We report the averaged F1-score of three coreference evaluation metrics: MUC, B$^3$, and CEAF, following~\citet{xu-choi-2020-revealing}.

\subsection{Language Understanding}
\paragraph{GLUE.} CoLA~\cite{warstadt2019neural} and SST-2~\cite{socher2013recursive_sst-2} are single-sentence tasks; MRPC~\cite{dolan2005automatically} and QQP are paraphrase detection tasks; MNLI~\cite{N18-1101}, QNLI~\cite{rajpurkar-etal-2016-squad}, and RTE~\cite{dagan2005pascal, haim2006second, giampiccolo2007third, giampiccolo2008fourth, bentivogli2009fifth} are inference tasks; STS-B~\cite{cer2017semeval} is a sentence similarity task. We report the accuracy for SST-2, MNLI, QNLI, RTE, and STS-B, and the Matthews correlation coefficient for CoLA. Both the accuracy and the F-1 score are included for MRPC and QQP.

We use the \texttt{run\_glue.py} script provided by HuggingFace \citep{wolf-etal-2020-transformers}, and follow their exact hyper-parameters. For all tasks, we use a batch size of 32, a maximum sequence length of 128, and a learning rate of $2 \times 10^{-5}$. We train for 3 epochs for all tasks except for MRPC, which is trained for 5.

\section{Standard Deviation}
\label{sec:appendix_std}
As many fairness benchmarks tend to exhibit high variance, we report the standard deviation across 3 runs for each of our implementations from the main results. Standard deviations for intrinsic tasks can be found in \autoref{apptab:ss}, and for extrinsic tasks in \autoref{apptab:ss2}. 


\begin{table}[t]
\centering
\setlength{\tabcolsep}{5pt} 
\resizebox{0.8\columnwidth}{!}{
\begin{tabular}{l cc c } 

\toprule
& \multicolumn{2}{c}{\tf{StereoSet}} & \tf{CrowS-Pairs} \\ 
\cmidrule(lr){2-3} \cmidrule(lr){4-4}  
\textbf{Model}   & \textbf{LM} & \textbf{SS} & \textbf{SS} \\ \midrule
\textsc{Context-Debias}             & {0.07}     & {0.24}        & {0.77} \\
\textsc{\ff{}}  & {1.47} & {0.71}  & {3.43} \\
 \midrule
\textsc{\ours{}} (ours)& {0.39}                & {0.69}                 & {0.77}     \\ 
\bottomrule
\end{tabular}}
\caption{Standard deviation on intrinsic tasks from StereoSet and CrowS-Pairs (\autoref{tab:ss}).}
\label{apptab:ss}
\end{table}

\begin{table*}[t]
\centering
\setlength{\tabcolsep}{5pt} 
\resizebox{\textwidth}{!}{
\begin{tabular}{l ccccc cccc ccccccc} 

\toprule
&  \multicolumn{5}{c}{\tf{Bias-in-Bios}} & \multicolumn{4}{c}{\tf{Bias-NLI}} & \multicolumn{7}{c}{\tf{WinoBias}} \\ 
\cmidrule(lr){2-6} \cmidrule(lr){7-10}  \cmidrule(lr){11-17}
\textbf{Model}   & \textbf{Acc. } & \textbf{Acc.} & \textbf{Acc. } & \textbf{TPR} & \textbf{TPR} & \textbf{TN} & \textbf{FN} & \textbf{T:0.5} & \textbf{T:0.7} & \textbf{ON} & \textbf{1A} & \textbf{1P} & \textbf{2A} & \textbf{2P} & \textbf{TPR-1} & \textbf{TPR-2}\\ 
& (All) & (M) & (F) & GAP & RMS &\\
\midrule
\textsc{BERT}    & {0.56} & {0.56} & {0.58} & {0.14} & {0.01} & {0.03} & {0.05} & {0.05} & {0.05} & {0.05} & {1.12} & {0.57} & {1.02} & {1.15} & {1.17} & {1.21}\\ \midrule
\textsc{Sent-Debias} & {0.09} & {0.10} & {0.20} & {0.25} & {0.00} & {0.05} & {0.04} & {0.05} & {0.09} & {0.15}  & {2.88} & {0.27} & {1.47} & {0.29} & {3.02} & {1.53}\\ 
\textsc{Context-Debias}            & {0.36} & {0.32} & {0.42} & {0.17} & {0.01} & {0.03} & {0.05} & {0.05} & {0.04} & {0.19} & {1.00} & {1.25} & {1.13} & {1.33} & {0.75} & {0.74}\\
\textsc{\ff{}}  & {0.07} & {0.25} & {0.16} & {0.41} & {0.00} & {0.05} & {0.07} & {0.01} & {0.05} & {0.64} & {0.41} & {0.47} & {2.31} & {1.49} & {0.16} & {0.84}\\
 \midrule
\textsc{\ours{}} (ours)                  & {0.40} & {0.51} & {0.47} & {0.04} & {0.00} & {0.02} & {0.01} & {0.01} & {0.03} &  {0.37} & {1.12} & {1.11} & {0.30} & {0.41} & {1.88} & {0.44}    \\ 
\bottomrule
\end{tabular}}
\caption{Standard deviation on extrinsic tasks: Bias-in-Bios (\autoref{tab:occ-cls}), Bias-NLI (\autoref{tab:nli}), and OntoNotes and WinoBias (\autoref{tab:coref}).}
\label{apptab:ss2}
\end{table*}

\section{SentEval Transfer Evaluation}
\label{sec:appendix_senteval}

To more thoroughly evaluate our models' capacity for NLU retention, we additionally test on 9 transfer tasks provided by the SentEval toolkit \cite{conneau2018senteval}. 

While some task overlap exists between SentEval and GLUE, the evaluation setup is different. By freezing the encoder and training a logistic regression classifier, the default SentEval implementation focuses on evaluating the knowledge stored in a fixed-size frozen sentence embedding. Whereas when testing on GLUE, the parameters in the entire encoder are allowed to freely update. GLUE also emphasizes high-resource downstream tasks, which use training data with hundreds of thousands of samples. Comparatively, SentEval mostly focuses on low-resource transfer, with smaller downstream classification tasks such as Movie Review (MR) or Product Review (CR) \cite{conneau2018senteval}. 

We use the evaluation toolkit provided by \citet{conneau2018senteval}, following the standard settings, and form sentence representations by extracting the \texttt{[CLS]} token of the last hidden state. 10-fold cross-validation was used for MR, CR, SUBJ, MPQA, and SST-2, and cross-validation for TREC. For MRPC, a 2-class classifier learns to predict the probability distribution of relatedness scores; the Pearson correlation coefficient is reported.

\autoref{tab:senteval} shows the performance across the downstream SentEval transfer tasks. As this evaluation regime does not involve fine-tuning, we notice less uniformity across the baselines' results. While \ours{} and \ours{} w/o MLM have a higher average performance than BERT, the only task with an obvious gain is in MRPC (68.87\% and 71.48\% vs. 65.57\%). This makes sense as MRPC is a semantic similarity task, which leveraging supervised signals from NLI entailment pairs happens to benefit \cite{gao2021simcse}.

\begin{table*}[t]
\centering
\resizebox{0.9\textwidth}{!}{
\begin{tabular}{lcccccccc} \toprule
\tf{Model}  & \tf{MR} $\uparrow$ & \tf{CR} $\uparrow$ & \tf{SUBJ} $\uparrow$     & \tf{MPQA} $\uparrow$      & \tf{SST-2} $\uparrow$ & \tf{TREC} $\uparrow$ & \tf{MRPC} $\uparrow$ & \tf{Avg.} $\uparrow$    \\  \midrule
\textsc{BERT}   & \textbf{80.99} & 85.67 & \textbf{95.31} & 87.40 & \textbf{86.99} & 84.20 & 65.57 & 83.73 \\ 
\textsc{Context-Debias}  & 78.37 & 85.22 & 94.11 & 85.99 & 84.73 & \tf{85.60} & 66.55 & 82.94 \\
\textsc{Sent-Debias} & 69.85 & 68.96 & 89.12 & 80.78 & 81.44 & 60.00 & 70.14 & 74.33 \\
\textsc{\ff{}} & 76.94 & 80.34 & 92.82 & 83.54 & 81.88 & 79.20 & 69.28 & 80.57\\ \midrule
\textsc{\ours{}} & 78.33 & 85.83 & 93.78 & \tf{89.13} & 85.50 & 85.20 & 68.87 & 83.81 \\
\textsc{\ours{} w/o MLM}  & 80.01 & \tf{86.41} & 94.50 & 89.29 & 85.45 & 84.80 & \tf{71.48} & \tf{84.56} \\ 
\bottomrule

\end{tabular}}
\caption{Downstream transfer task results for BERT, \ours{} models, and bias baselines from the SentEval benchmark \cite{conneau-etal-2017-supervised}.}
\label{tab:senteval}
\end{table*}

\section{Bias Benchmarks in Other Works}
\label{sec:appendix_benchmarks}
\begin{table*}[ht]
\setlength{\tabcolsep}{3pt}
 \centering 
    \resizebox{\textwidth}{!}{
\begin{tabular}{@{}lccccccll@{}}
\toprule
& \tf{WEAT/} & \tf{CrowS-} & \tf{Stereo-} & \tf{Bias-} & \tf{Bias-} & \tf{Wino-} & & \\
\tf{Method} & \tf{SEAT} & \tf{Pairs} & \tf{Set} & \tf{in-Bios} & \tf{NLI} & \tf{Bias} & \tf{Other int.} & \tf{Other ext.} \\ \midrule
\textbf{Task-specific approaches}  \\
~~\textsc{INLP} \cite{ravfogel-etal-2020-null} &   &   &   &  \cmark${^\ast}$ &   &   &   &   \\
~~\textsc{R-LACE}   \cite{ravfogel2022linear} &   &   &   &  \cmark${^\ast}$ &   &   &   &   \\
~~\textsc{Con} \cite{shen-2021-contrastive} &   &   &   &  \cmark${^\ast}$ &   &   &   &   \\
~~\textsc{DADV}  \cite{han2021diverse} &   &   &   &  \cmark${^\ast}$ &   &   &   &   \\
~~\textsc{GATE}  \cite{han2021balancing} &   &   &   &  \cmark${^\ast}$ &   &   &   &   \\ \midrule 
\textbf{Task-agnostic approaches} \\
~~\textsc{CDA} \cite{webster2020measuring} &   &   &   &  \cmark &   &   & DisCo & STS-B, WinoGender \\
~~\textsc{Dropout} \cite{webster2020measuring} &   &   &   &  \cmark &   &   & DisCo & STS-B, WinoGender \\
~~\textsc{ADELE}  \cite{lauscher-etal-2021-sustainable-modular} &  \cmark &   &   &   &  \cmark &   & BEC-Pro, DisCo & STS-B \\
~~\textsc{Bias Projection} \cite{Dev_Li_Phillips_Srikumar_2020}&   &   &   &   &  \cmark &   &   &   \\
~~\textsc{OSCaR} \cite{dev2021oscar} &  \cmark &   &   &   &  \cmark &   & ECT  &  SIRT \\
~~\textsc{Sent-Debias} \cite{liang-etal-2020-towards}  &  \cmark &   &   &   &   &   &   &   \\
~~\textsc{Context-Debias} \cite{kaneko-bollegala-2021-context} &  \cmark &   &   &   &  \cmark &   &   &   \\
~~\textsc{Auto-Debias} \cite{guo-etal-2022-auto} & \cmark &  \cmark${^\ast}$  &   &   &  &   &   &   \\
~~\textsc{\ff{}}   \cite{ChengHYSC21}  &  \cmark &   &   &   &   &   &   &   \\
~~\textsc{\ours{}} (ours)  &   &  \cmark${^\ast}$ &  \cmark${^\ast}$ &  \cmark${^\ast}$ &  \cmark${^\ast}$ &  \cmark${^\ast}$ &      \\ \bottomrule
\end{tabular}}
\caption{Gender bias metrics used for each baseline, as reported from the original work. A $^\ast$ means that the metrics are directly comparable to those in our main results. DisCo \citep{webster2020measuring} is a template-based likelihood metric; STS-B is a semantic similarity task adapted by \citet{webster2020measuring} for measuring gender bias; WinoGender \cite{rudinger-EtAl:2018:N18} is a small-scale coreference resolution dataset that links pronouns and occupations; BEC-Pro \cite{bartl2020unmasking} is a template-based metric that measures the influence of an occupation word on a gender word; ECT \cite{pmlr-v89-dev19a} applies the Spearman's correlation coefficient to calculate the association between gender words and attribute-neutral words, SIRT \cite{dev2021oscar} uses NLI data to evaluate for gendered information retention.}
\label{tab:baseline_metrics_chart}
\end{table*}
\autoref{tab:baseline_metrics_chart} shows a comprehensive compilation of gender bias metrics used by other bias mitigation methods in recent literature.
\section{SEAT Evaluation}
\label{sec:appendix_seat}

The Sentence Encoder Association Test (SEAT) \cite{may-etal-2019-measuring} is the sentence-level extension of the Word Embedding Association Test (WEAT) \cite{caliskan2017semantics}, a hypothesis-driven diagnostic that checks whether two sets of target words (for instance, \texttt{[artist, musician,...]} and \texttt{[scientist, engineer...]}) are equally similar to two sets of attribute words (for instance, \texttt{[man, father,...]} and \texttt{[woman, mother,...]}). In SEAT, the concept words from WEAT are inserted into semantically bleached sentence templates such as ``This is a[n] $<$word$>$," effectively allowing for the comparison of \textit{sentence} representations. 

For both WEAT and SEAT, the null hypothesis postulates that no difference exists in the relative similarity between the sets of target words $X$, $Y$ and the sets of attribute words $A$, $B$. The effect size, $s(X,Y,A,B)$, quantifies the difference in mean cosine similarity between representations of the target concept pair ($X$, $Y$), and representations of the attribute concept pair ($A$, $B$), through the following equations:
\begin{align*}
    & \scalebox{0.9}{$s(w, A, B) = \underset{a \in A}{\textrm{mean}} \cos(\overrightarrow{w}, \overrightarrow{a}) - \underset{b \in B}{\textrm{mean}}  \cos(\overrightarrow{w}, \overrightarrow{b})$} \\
    & \scalebox{0.9}{$s(X, Y, A, B) = \sum_{x \in X} s(x, A, B) - \sum_{y \in Y} s(y, A, B)$}
\end{align*}

We report our results (alongside pre-trained BERT's) in \autoref{tab:app_seat_mabel}. Following \citet{liang-etal-2020-towards}, we extract the sentence representation as the \texttt{[CLS]} token fed through a linear layer and \texttt{tanh} activation (e.g., the pooled output). 

\begin{table}[ht]
\begin{tabular}{@{}lcc@{}}
\toprule 
\multicolumn{1}{l}{\tf{SEAT}} & \multicolumn{1}{l}{\tf{BERT}} & \multicolumn{1}{l}{\tf{\ours{}}}   \\
\tf{Category} &  \tf{ES} $\ddagger$ & \tf{ES} $\ddagger$  \\ \midrule
Names, Career/Family 6  & 0.477 & 0.664\small{±0.313}  \\
Terms, Career/Family 6b  & 0.108 & 0.167\small{±0.196}  \\
Terms, Math/Arts 7  & 0.253 & 0.479\small{±0.488}   \\
Names, Math/Arts 7b  & 0.254 & 0.647\small{±0.254}   \\
Terms, Science/Arts 8  & 0.399 & 0.465\small{±0.288}   \\
Names, Science/Arts 8b  & 0.636 & 0.570\small{±0.296}   \\
Avg. Abs. Effect Size & 0.354 & 0.499\small{±0.090}  \\ \bottomrule
\end{tabular}
\caption{BERT's and \ours{} effect sizes (ES) on the gender-associated SEAT categories. For \ours{}, we report the absolute average and standard deviation across 3 runs. $\ddagger$: the closer to 0, the better.}
\label{tab:app_seat_mabel}
\end{table}
\section{A Comparison to SimCSE}
\label{sec:appendix_simcse}
 A non-debiasing analogue to \ours{} is supervised SimCSE \cite{gao2021simcse}, a state-of-the-art representation learning approach that generates sentence representations with good semantic textual similarity (STS) performance. Like \ours{}, supervised SimCSE also trains on entailment pairs from NLI data using an contrastive learning objective. 

One potential concern is that \ours{} performs well on tasks such as Bias-NLI not due to greater fairness, but because it has already been trained on NLI data. To test this assumption, we repeat the Bias-NLI task on SimCSE, which has been trained on un-augmented entailment pairs from the SNLI dataset. In \autoref{tab:simcse_nli}, the tangible increase across all metrics from SimCSE to \ours{} indicates that \ours{}'s good performance can not solely be attributed to NLI knowledge retention.

\begin{table}[t]
\centering
\resizebox{\columnwidth}{!}{%
\begin{tabular}{lcccc} \toprule
      \tf{Model} & \tf{TN}$\uparrow$ & \tf{FN}$\uparrow$ & \tf{T:0.5} $\uparrow$ & \tf{T:0.7}  $\uparrow$ \\ \midrule
\textsc{Sup. SimCSE} & 0.830 & 0.951 & 0.945 & 0.839  \\ 
\textsc{\ours{}} (ours) & \bf{0.917} & \bf{0.983} & \bf{0.983} & \bf{0.968}\\\bottomrule
\end{tabular}}
\caption{Results on Bias-NLI for supervised SimCSE \cite{gao2021simcse} and \ours{}.} 
\label{tab:simcse_nli}
\end{table}
\section{Linear Probing Experiments}
\label{sec:appendix_probing}

To better illustrate the effects of \ours{}, we conduct a suite of probing experiments, in which the entire \textit{encoder} parameters are frozen during training. We summarize task details and results below. 

\subsection{Bias-in-Bios}
We follow the same procedure as in the fine-tuning setting, except freeze the encoder and only update the linear classification layer. From \autoref{tab:occ-cls-fixed}, both \ours{} and \ours{} without the MLM loss show better overall and gender-specific task accuracy on the probe, in comparison to BERT and other bias mitigation baselines. \ours{} has the lowest TPR-GAP and the second lowest TPR-RMS. While INLP has a very low TPR-RMS of 0.069, it also has significantly worse accuracy, whereas \ours{} achieves a better fairness-accuracy balance. 

\begin{table}[t]
\centering
\resizebox{0.5\textwidth}{!}{
\begin{tabular}{lccccc} \toprule
      & \tf{Acc.}  & \tf{Acc.} & \tf{Acc.} & \tf{TPR} & \tf{TPR} \\ 
      \tf{Model} & \tf{(All)} $\uparrow$ & \tf{(M)} $\uparrow$ & \tf{(F)} $\uparrow$ & \tf{GAP} $\downarrow$ & \tf{RMS} $\downarrow$ \\\midrule
\multicolumn{6}{c}{\textit{Bias-in-Bios - Linear Probe}} \\ \midrule
\textsc{BERT}   & 79.63        & 80.27    & 78.84    & 1.436   & 0.200   \\
\textsc{Context-Debias}   & 78.27        & 78.98    & 77.39    & 1.595   & 0.214   \\
\textsc{Sent-Debias} & 75.55        & 76.19    & 74.74    & 1.452   & 0.195   \\
\textsc{\ff{}} & 74.69        & 75.30    & 73.94    & 1.357   & 0.225   \\
\textsc{INLP}   & 72.36        & 73.36   & 71.10    & 2.261   & \bf{0.069}   \\ \midrule
\textsc{\ours{} w/o  MLM}  & 80.68        & 81.22    & 80.00    & 1.220   & 0.172   \\ 
\textsc{\ours{}}  & \bf{80.98}       & \bf{81.43}    & \bf{80.41}    & \bf{1.012}   & 0.159 \\
\bottomrule
\end{tabular}}
\caption{Linear probing results on Bias-in-Bios across the \ours{} models and different baselines. }
\label{tab:occ-cls-fixed}
\end{table}

\subsection{Bias-NLI}

\citet{Dev_Li_Phillips_Srikumar_2020} originally formulated this task as a probing experiment. Accordingly, we only update a linear layer on top of a frozen encoder when training on the SNLI dataset. We freeze the entire model when evaluating on the test dataset. Our results are shown in \autoref{tab:res_nli_fixed}.

\begin{table}[t]
\resizebox{1.0\columnwidth}{!}{%
\begin{tabular}{lccc} \toprule
    \tf{Model}  & \tf{TN} $\uparrow$ & \tf{FN} $\uparrow$ & \tf{T:0.5} $\uparrow$ \\ \midrule
    \multicolumn{4}{c}{\textit{Bias-NLI - Linear Probe}} \\ \midrule
$\textsc{BERT}^{\star}$                 & 0.409 & 0.512 & 0.239 \\
\textsc{Sup. SimCSE}               & 0.502 & 0.792 & 0.516 \\ 
\textsc{Sup. SimCSE + MLM} & 0.412 & 0.551 & 0.264 \\ \midrule
\textsc{Bias Projection}$^{\star}$ - Test         & 0.396 & 0.371 & 0.341 \\ 
\textsc{Bias Projection}$^{\star}$- Train + Test            & 0.516 & 0.526 & 0.501 \\ 
\textsc{OSCaR}$^{\star \dagger}$ & 0.566 & 0.588 & -  \\ 
\textsc{Sent-Debias}               & 0.351 & 0.319 & 0.020 \\
\textsc{Context-Debias}               & 0.240 & 0.078 & 0.023 \\ 
\textsc{\ff{}}               & 0.348 & 0.318 & 0.055 \\\midrule
\textsc{MABEL w/o MLM} & \tf{0.571} & \tf{0.853} & \tf{0.710} \\
\textsc{MABEL} & 0.538 & 0.837 & 0.653 \\
\bottomrule
\end{tabular}}
\caption{Natural language inference results for pre-trained \textsc{BERT}, the baselines, and \textsc{MABEL}. Best numbers in \textbf{bold}. $\textsc{w/o}$ = without; $\star$: results are from original papers; $\dagger$: the encoder model is RoBERTa$_{\text{base}}$; $\ddagger$: the models are fine-tuned on MNLI. \textsc{BERT} and \textsc{Bias Projection} results are from \citet{Dev_Li_Phillips_Srikumar_2020}; \textsc{OSCaR} is from \citet{dev2021oscar}.} 
\label{tab:res_nli_fixed}
\end{table}

\begin{table*}[ht]
\centering
\resizebox{0.45\textwidth}{!}{
\begin{tabular}{@{}lcccc@{}}
\toprule
 \tf{Metric} & \multicolumn{1}{l}{\textbf{BERT}} & \multicolumn{1}{l}{\textbf{AL1}} &  \multicolumn{1}{l}{\textbf{AL2}} & \multicolumn{1}{l}{\textbf{AL3}} \\ \midrule
 \multicolumn{5}{c}{\textit{Bias-in-Bios - Linear Probe}} \\
\midrule
Overall Acc. ↑ & 79.63 & \tf{80.68} &  79.94 & 75.30  \\
Acc. (M) ↑ & 80.27 & \tf{81.22} & 80.63 & 77.32  \\
Acc. (F) ↑ & 78.83 & \tf{80.00}  & 79.08 & 72.77  \\
TPR GAP  ↓ & 1.436 & \tf{1.220}  & 1.550 & 4.543  \\
TPR RMS ↓ & 0.200 & \tf{0.172}  & 0.180 & 0.234  \\
\midrule
\multicolumn{5}{c}{\textit{Bias-NLI - Linear Probe}} \\
\midrule
NN ↑ & 0.409 & \textbf{0.571} & 0.509 & 0.354  \\
FN ↑ & 0.512 & \textbf{0.853} & 0.737 & 0.318 \\
T:0.5 ↑ & 0.239 & \textbf{0.710}  & 0.538 & 0.069 \\
\midrule
\multicolumn{5}{c}{\textit{Bias-NLI - Fine-tuning}} \\
\midrule
NN ↑ & 0.762 & \tf{0.917}   & 0.897 & 0.844 \\
FN ↑ & 0.900 & \tf{0.983}   & 0.948 & 0.927 \\
T:0.5 ↑ & 0.718 & \tf{0.983}   & 0.949 & 0.920 \\ \bottomrule
\end{tabular}
}
\centering \caption{Results on Bias-in-Bios (linear probe setting) and Bias-NLI (linear probe setting and fine-tuning setting), for \ours{} trained with different alignment losses---AL1 (default), AL2, and AL3. All models are trained only on SNLI data in this ablation study.}
\label{tab:align}
\end{table*}

\begin{figure}
\centering
\includegraphics[scale=0.48]{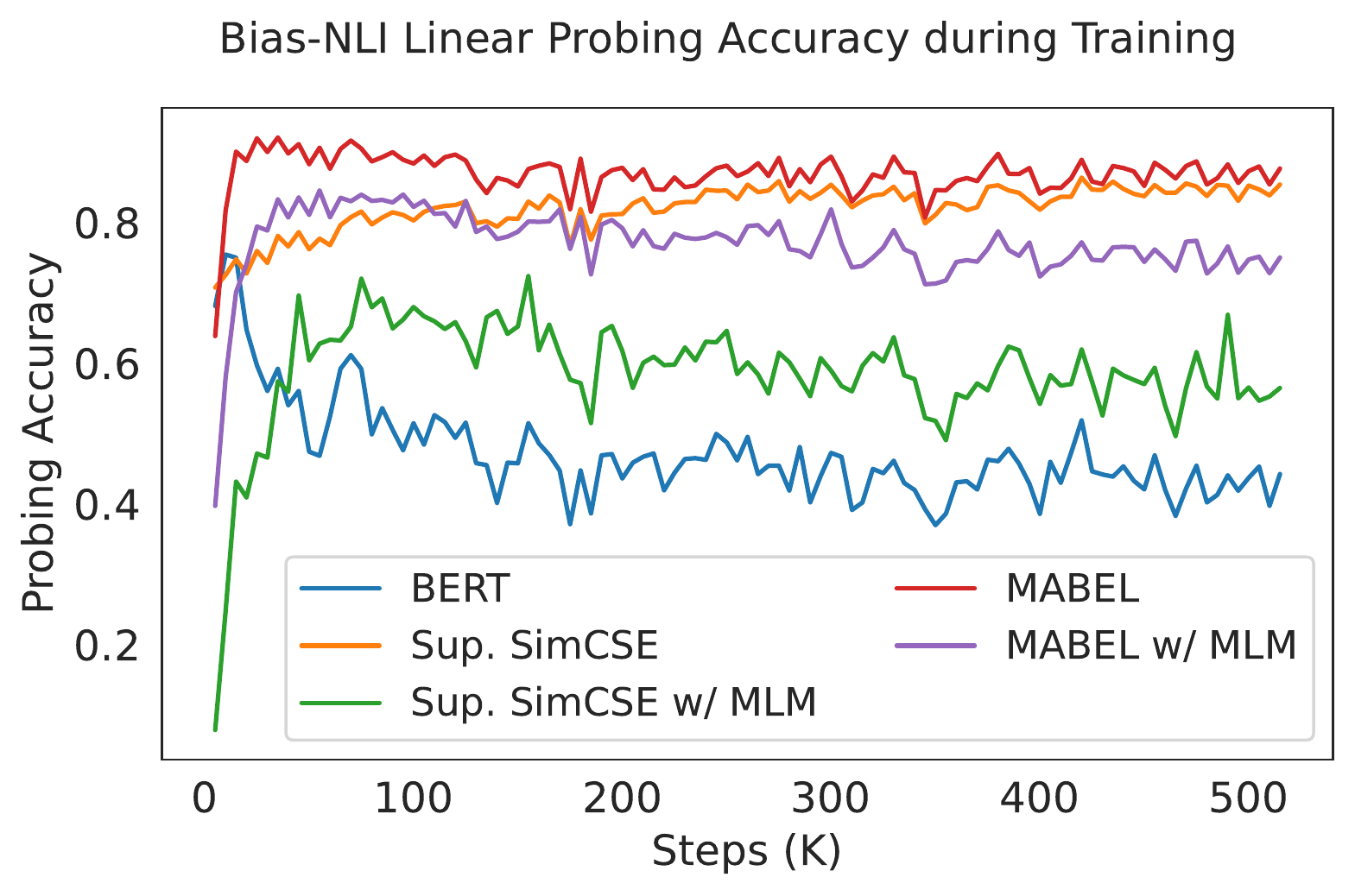}
\caption{Linear probing accuracy on Bias-NLI of model checkpoints at various timesteps when training on SNLI.}
\label{fig:nli-probe}
\end{figure}

\autoref{fig:nli-probe} shows the validation accuracy of BERT, \ours{} models, and supervised SimCSE models on the NLI-Bias evaluation set at various timesteps when training on the SNLI dataset. BERT consistently struggles with the task---its accuracy starts off high before degrading noticeably. \ours{} outperforms SimCSE both with and without the MLM loss, which shows that its performance on Bias-NLI is not entirely due to enhanced semantic understanding, but greater fairness as well. The MLM objective steadily drops the performance of both \ours{} and SimCSE, which shows that it harms sentence-level knowledge retention. Interestingly, the opposite trend holds true in the fine-tuning setting---including the MLM loss leads to better NLI performance across all three metrics.


\section{Alignment Objectives}
\label{sec:appendix_align}

Beside our default alignment loss (\tf{Alignment Loss 1}), we experiment with other losses that maximize the similarity between original and gender-augmented representations. We describe them and report their results across Bias-in-Bios and Bias-NLI tasks.

\paragraph{Alignment Loss 2} is a contrastive objective that is similar to \ff{}, but takes cosine similarity as a scoring function instead of a two-layer fully-connected neural network. Augmented sentence pairs, either $(p,p')$ (or $(h,h')$), form positive pairs, and other sentences form in-batch negatives. Let $x_i$ be any original premise or hypothesis representation, and $x'$ be the augmented counterpart of $x$:

\begin{equation*}
    \mathcal{L}_{AL2} = - \log{\frac{e^{\textrm{sim}(x_i, x_i') / \tau}}{\sum_{j=1}^{2n} e^{\textrm{sim}(x_i, x_j)/ \tau} }}.
\end{equation*}


\paragraph{Alignment Loss 3} tries to maximize the cosine similarities between original and augmented sentences. Given the pairs $(p_i,p_i')$ and $(h_i, h_i')$:

\begin{equation*}
    \mathcal{L}_{AL3} = \frac{1}{n} \sum_{i=1}^n -(\textrm{sim}(p_i, p'_i) - \textrm{sim}(h_i, h'_i)).
\end{equation*}







The results for \ours{} trained with each alignment loss are in \autoref{tab:align}. Our default alignment loss (AL1) returns consistently better results.

\end{document}